\documentclass[11pt]{article}

\usepackage[final]{acl}

\usepackage{times}
\usepackage{latexsym}

\usepackage[T1]{fontenc}

\usepackage[utf8]{inputenc}

\usepackage{microtype}

\usepackage{inconsolata}

\usepackage{graphicx}
\usepackage{multirow}
\usepackage{booktabs}
\usepackage{multicol}
\usepackage{array}
\usepackage{threeparttable}
\usepackage{amssymb} 
\usepackage[most]{tcolorbox}
\usepackage{minted}
\fvset{breaklines=true}

\usepackage{subcaption} 
\usepackage{listings}
\usepackage{algorithm}
\usepackage[noend]{algorithmic}

\usepackage[table]{xcolor}
\usepackage{pgfplotstable}

\usepackage{pifont} 

\newcolumntype{C}[1]{>{\centering\arraybackslash}p{#1}}

\definecolor{systemcolor}{rgb}{0.9,0.9,1} %
\definecolor{usercolor}{rgb}{1,0.9,0.9} %

\newcommand{\fmlsymbol}[0]{\text{\smash{\raisebox{-1.5pt}{\includegraphics[height=9pt]{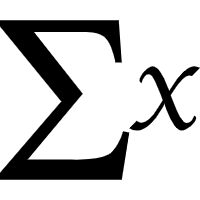}}}}}

\newcommand{\imgsymbol}[0]{\text{\smash{\raisebox{-1.5pt}{\includegraphics[height=9pt]{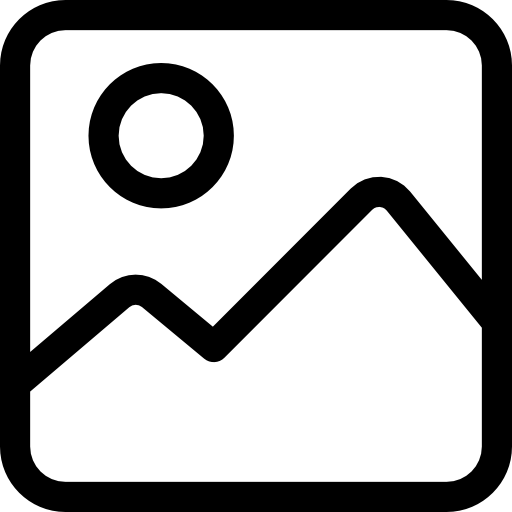}}}}}

\newcommand{\textsymbol}[0]{\text{\smash{\raisebox{-1.5pt}{\includegraphics[height=9pt]{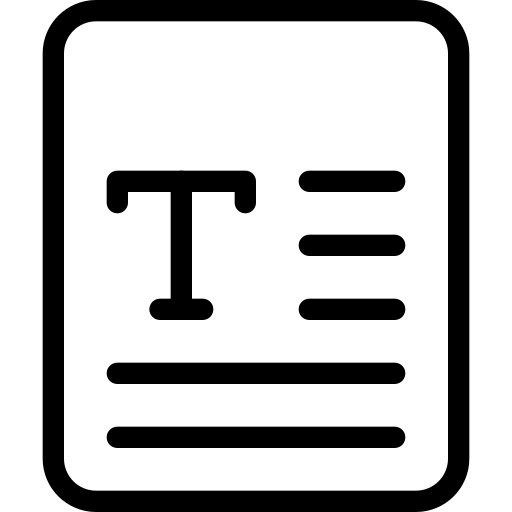}}}}}
\newcommand{\tabsymbol}[0]{\text{\smash{\raisebox{-1.5pt}{\includegraphics[height=9pt]{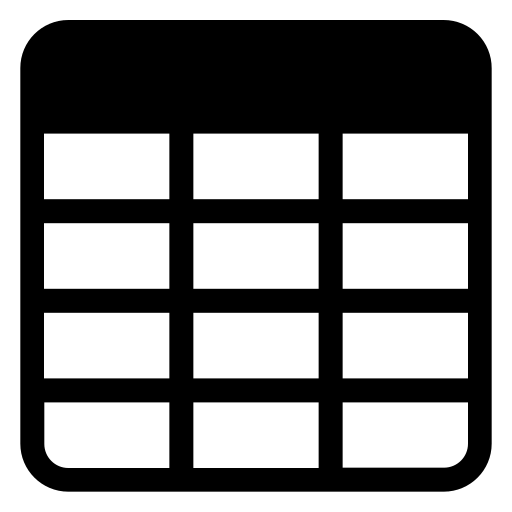}}}}}

\definecolor{myDarkGreen}{RGB}{0, 100, 0}
\newcommand{\cmark}
{\textcolor{myDarkGreen}{\ding{51}}}%
\newcommand{\xmark}{\textcolor{red}{\ding{55}}}%

\pgfplotstableset{
color cells/.style={
    col sep=comma,
    string type,
    postproc cell content/.code={%
            \pgfkeysalso{@cell content=\rule{0cm}{2.4ex}%
            \pgfmathsetmacro\y{min(100,max(0,abs(round(##1 * 0.5))))}%
            \ifnum##1<0\edef\temp{\noexpand\cellcolor{blue!\y}}\temp\fi%
            \ifnum##1>0\edef\temp{\noexpand\cellcolor{red!\y}}\temp\fi%
            \pgfmathtruncatemacro\x\y%
            \ifnum\x>50 \color{white}\fi%
            ##1}%
            }
}
}

\usepackage{adjustbox}
\usepackage{tcolorbox}

\newtcolorbox{combinedprompt}{
    colframe=black,
    colback=white,
    boxrule=0.6mm,
    width=\linewidth,
    fonttitle=\bfseries,
    rounded corners,
    coltitle=black
}

\usepackage{listings} %
\lstnewenvironment{configpython}[2][] %
{
    \lstset{
        commentstyle=\color{UBlau},
        keywordstyle=\color{WiWi},
        numberstyle=\tiny\color{Grau60},
        basicstyle=\ttfamily\footnotesize,
        language=python,
        numbers=none,
        frame=tblr,           %
        tabsize=4,
        captionpos=b,         %
        caption={#1},
        label={#2},
        breaklines=true,      %
        breakatwhitespace=false,
        breakautoindent=true,
        prebreak=\mbox{\textcolor{red}{$\hookrightarrow$}} %
    }
}{}

\lstnewenvironment{configjson}[2][] %
{
    \lstset{
        commentstyle=\color{UBlau},
        keywordstyle=\color{WiWi},
        numberstyle=\tiny\color{Grau60},
        basicstyle=\ttfamily\footnotesize,
        numbers=none,
        frame=tblr,           %
        tabsize=4,
        captionpos=b,         %
        caption={#1},
        label={#2},
        breaklines=true,      %
        breakatwhitespace=false,
        breakautoindent=true,
        prebreak=\mbox{\textcolor{red}{$\hookrightarrow$}} %
    }
}{}

\usepackage{amssymb}
\usepackage{amsmath}
\usepackage{float} 
\title{MMKG-RDS: Reasoning Data Synthesis via Deep Mining of Multimodal Knowledge Graphs}

\author{Lun Zhan$^{*}$, Feng Xiong$^{*}$, Huanyong Liu$^{\dagger}$, Feng Zhang, Yuhui Yin\\
  AI Research Institute, Qihoo 360\\
  {\tt\small zhanlun@360.cn}, {\tt\small teddybear@tju.edu.cn}, {\tt\small liuhuanyong@360.cn} \\{\tt\small xhpzrs850704@gmail.com}, {\tt\small yinyuhui.ares@foxmail.com}
}

\begin{document}
\maketitle

\renewcommand{\thefootnote}{\fnsymbol{footnote}}  
\footnotetext[1]{These authors contributed equally.}
\footnotetext[2]{Corresponding author.}

\begin{abstract}
Synthesizing high-quality training data is crucial for enhancing domain models' reasoning abilities. Existing methods face limitations in long-tail knowledge coverage, effectiveness verification, and interpretability. Knowledge-graph-based approaches still fall short in functionality, granularity, customizability, and evaluation. To address these issues, we propose MMKG-RDS—a flexible framework for reasoning data synthesis that leverages multimodal knowledge graphs. It supports fine-grained knowledge extraction, customizable path sampling, and multidimensional data quality scoring. We validate MMKG-RDS with the MMKG-RDS-Bench dataset, covering five domains, 17 task types, and 14,950 samples. Experimental results show fine-tuning Qwen3 models (0.6B/8B/32B) on a small number of synthesized samples improves reasoning accuracy by 9.2\%. The framework also generates distinct data, challenging existing models on tasks involving tables and formulas, useful for complex benchmark construction. 
The dataset and code are available at \url{https://github.com/360AILAB-NLP/MMKG-RDS}
\end{abstract}
\section{Introduction}
High-quality training data~\citep{2305.11206} is essential for improving domain-specific model reasoning. Automated data synthesis~\citep{2304.12244, 2212.10560} has become key to overcoming data bottlenecks. However, current methods face three challenges: \textbf{insufficient long-tail knowledge coverage}~\citep{dong2024once, 2410.17355, 2311.07237}, leading to poor performance in rare cases; \textbf{difficulty in effectiveness verification}~\citep{2309.07601}, where automatic evaluation signals misalign with true performance; and \textbf{lack of interpretability}~\citep{2301.13379, 2305.07722}, hindering quality control.

Recent advancements in knowledge graphs have highlighted their semantic advantages, yet existing synthesis frameworks exhibit significant limitations. Methods like Deepdive~\citep{2509.10446} and MedResearcher-R1~\citep{2508.14880} rely on external graphs, lacking end-to-end document parsing and graph construction. Tools such as DataFlow~\citep{2512.16676} fail to fully exploit graph structures. Regarding knowledge representation, frameworks like GraphGen~\citep{2505.20416}, Deepdive~\citep{2509.10446}, and GRIP~\citep{2412.08864} are limited to homogeneous graphs in text, failing to capture multimodal information such as tables, formulas, and charts. Existing frameworks also lack support for flexible custom schemas, relying on coarse-grained ontologies or schema-free approaches, which limits their domain-specific applicability. Furthermore, they cover a narrow range of tasks and lack scalability mechanisms. For data analysis, current solutions provide only single-dimensional quality filtering, restricting deeper evaluation.

To address these issues, we propose MMKG-RDS, a framework for reasoning-based data synthesis, leveraging deep multimodal knowledge graphs. MMKG-RDS supports fine-grained knowledge extraction, customizable path sampling, and multidimensional quality scoring, addressing limitations in functionality, representation, customizability, and quality analysis.

The core advantages of MMKG-RDS include:
\begin{itemize}
\item \textbf{End-to-end automated graph construction}: Fully automates the process from document parsing to graph construction and quality filtering, compatible with pre-built graphs, enhancing domain adaptability.
\item \textbf{Deep multimodal knowledge representation}: Utilizes a heterogeneous graph with seven node types (Document, Chunk, Entity, Assertion, Image, Table, Formula) and 16 relationship types (see Table~\ref{tab:edge_domain_cnt}), capturing document structure and deep logic.
\item \textbf{Customizable mechanism}: Supports configurable domain schemas, dynamic task expansion (32+ types, see Table~\ref{tab:qa_task}), and precise control over node types, relationships, and sampling strategies.
\item \textbf{Multidimensional quality assessment}: Introduces evaluation metrics covering complexity, fidelity, and difficulty, offering comprehensive quality labels for synthesized data.
\item \textbf{Modular architecture and integration}: Features a pluggable design, compatible with large model APIs, Neo4j graph databases, and standardized benchmark datasets for fair comparisons.
\end{itemize}

The main contributions of this paper are as follows:
\begin{enumerate}
\item We propose MMKG-RDS, a flexible framework for reasoning-based data synthesis using multimodal knowledge graphs, enabling fine-grained knowledge extraction, customizable path sampling, and data quality scoring.
\item We construct MMKG-RDS-Bench, a complex reasoning dataset, and show through experiments that synthesized data improves model performance on cross-modal reasoning tasks, validating the effectiveness of our difficulty and diversity filtering modules.
\item We identify that tasks involving tables and formulas pose greater challenges for cross-modal reasoning, offering insights for future large-model reasoning benchmarks.
\end{enumerate}

\section{Related Work}

\begin{table*}[tb!] 
\centering
\footnotesize
\setlength\tabcolsep{1pt}
\vspace{-3mm}

\resizebox{\textwidth}{!}{

\begin{threeparttable}
\begin{tabular}{l|*{4}{C{1.1cm}} | *{2}{C{1.1cm}} C{1.3cm}| *{2}{C{1.2cm}} | *{3}{C{1.2cm}} }
\toprule
\multirow{3}{*}{\textbf{Framework}} & \multicolumn{4}{c}{Function} & 
\multicolumn{3}{c}{KG Data} & \multicolumn{2}{c}{User Design} & 
\multicolumn{3}{c}{Analysis}
\\ \cmidrule(l){2-13} 
& Doc Parse & KG Builder & Path Sample &  QA Filter
& {Node Type} & {Relation Type} & \multirow{2}{*}{Modal} 
& \multirow{2}{*}{Schema} & \multirow{2}{*}{Task} 
& \multirow{2}{*}{Dim} & {Statistic Display} & \multirow{2}{*}{Dataset}
\\ \midrule
 

Easy Dataset~\citep{2507.04009} & \cmark & \xmark & \xmark & \cmark & \xmark & \xmark & \textsymbol & \xmark & 1 & 1 & \xmark & \xmark \\
DataFlow~\citep{2512.16676} & \cmark & \xmark & \xmark & \cmark & \xmark & \xmark & \textsymbol & \xmark & 1 & 6 & \xmark & \xmark  \\
GraphGen~\citep{2505.20416} & \cmark & \cmark & \cmark & \cmark & 1 & 1 & \textsymbol & \xmark & 3 & 3 & \xmark & \xmark  \\
Deepdive~\citep{2509.10446} & \xmark & \xmark & \cmark & \cmark & 1 & 1 & \textsymbol & \xmark & 1 & 1 & \xmark & \xmark  \\
Medreason~\citep{2504.00993} & \xmark & \xmark & \cmark & \cmark & 1 & 1 & \textsymbol & \xmark & 1 & 1 & \xmark & \cmark  \\
MedResearcher-R1~\citep{2508.14880} & \cmark & \cmark & \cmark & \cmark & 1 & 1 & \textsymbol & \xmark & 1 & 1 & \xmark & \xmark  \\
Graph2Eval~\citep{2510.00507} & \cmark & \cmark & \cmark & \cmark & 6 & 9 & \textsymbol\imgsymbol\tabsymbol & \xmark & 30 & 3 & \xmark & \cmark  \\
UniDoc-Bench~\citep{2510.03663} & \cmark & \cmark & \xmark & \cmark & 1 & 1 & \textsymbol\imgsymbol\tabsymbol & \xmark & 4 & 5 & \xmark & \cmark  \\
Bottom~\citep{2507.13966} & \xmark & \xmark & \xmark & \cmark & 1 & 1 & \textsymbol & \xmark & 1 & 1 & \xmark & \cmark  \\
GRIP~\citep{2412.08864} & \cmark & \cmark & \xmark & \cmark & 1 & 1 & \textsymbol & \xmark & 1 & 1 & \xmark & \cmark  \\

\midrule
 MMKG-RDS (ours)    
 & \cmark
 & \cmark
 & \cmark
 & \cmark
 & 7
 & 16
 & \textsymbol \imgsymbol \tabsymbol \fmlsymbol
 & \cmark
 & 32+
 & 7
 & \cmark
 & \cmark \\ 

\bottomrule       
\end{tabular}
\begin{tablenotes}[flushleft]\scriptsize
\item \textbf{User Design}: The Schema defines the ontology and other related information for knowledge graph construction. The Task module in our framework predefines 32 types of relatively common tasks; in addition, the \textbf{+} indicates that users can generate customized tasks by combining various strategies via modifying configuration files.
\item \textbf{Analysis}: Dim denotes the number of dimensions that support statistical analysis, and we enable the analysis of generated data across 7 dimensions: token\_len, task\_type, domain, difficulty, complexity, support, and hop\_num.
\end{tablenotes}
\end{threeparttable}

}
\caption{Comparison of existing Framework with MMKG-RDS. }
\label{tab:comparison}
\vspace{-1mm}
\end{table*}
\subsection{Data Synthesis Framework}
Several approaches address data synthesis, as summarized in Table~\ref{tab:comparison}. In general frameworks,~\citet{2507.04009} generates question-answer pairs via GUI-configured role prompts, while~\citet{2512.16676} offers modular APIs for system-level data synthesis. Knowledge-graph-guided frameworks include~\citet{2505.20416}, which builds fine-grained knowledge graphs to detect knowledge gaps and generate diverse question-answer pairs, and~\citet{2412.08864}, which uses implicit relationships in KGs for efficient data expansion. Domain-specific frameworks focus on medical KGs, with ~\citet{2504.00993, 2508.14880, 2507.13966} generating multi-hop question-answer pairs, and~\citet{2509.10446} synthesizing complex data for deep search tasks. Multimodal frameworks include~\citet{2510.00507}, which uses multi-source data KGs for multimodal document understanding, and~\citet{2510.03663}, which extracts multimodal information for question-answer pair generation.

\subsection{Multimodal Document Parsing}
Tools like~\citet{2409.18839, 2506.05218, 2507.05595, 2408.09869} integrate OCR and layout analysis to convert documents into structured formats such as Markdown. Alternatively, Qwen2.5-VL-7B~\citep{2502.13923} employs a unified neural network architecture that processes entire pages directly, avoiding modular pipeline errors.

\subsection{Knowledge Graph Construction}
In knowledge graph construction,~\citet{2404.16130, 2410.05779, 2409.13731, 2510.12323} utilize predefined coarse-grained ontologies and LLMs to extract entities and relationships. In contrast, ~\citet{2510.09156} autonomously extends the ontology with real-time knowledge, while ~\citet{2505.23628} uses LLMs to extract entity-event triplets, avoiding manual ontology creation.

\subsection{Data Quality Scoring}
For data quality evaluation,~\citet{2312.15685} filters data using a dual-scoring model for complexity and quality.~\citet{2311.15653} applies LLMs to evaluate datasets across quality, coverage, and necessity.~\citet{2308.12032} proposes a self-guided metric to identify high-quality samples, while LLM-as-judge methods~\citep{2306.05685} directly score data quality, and clustering selection strategies~\citep{2305.09246} enhance diversity through embedding clustering.

\begin{figure*}[t]
    \centering
    \includegraphics[width=1\linewidth]{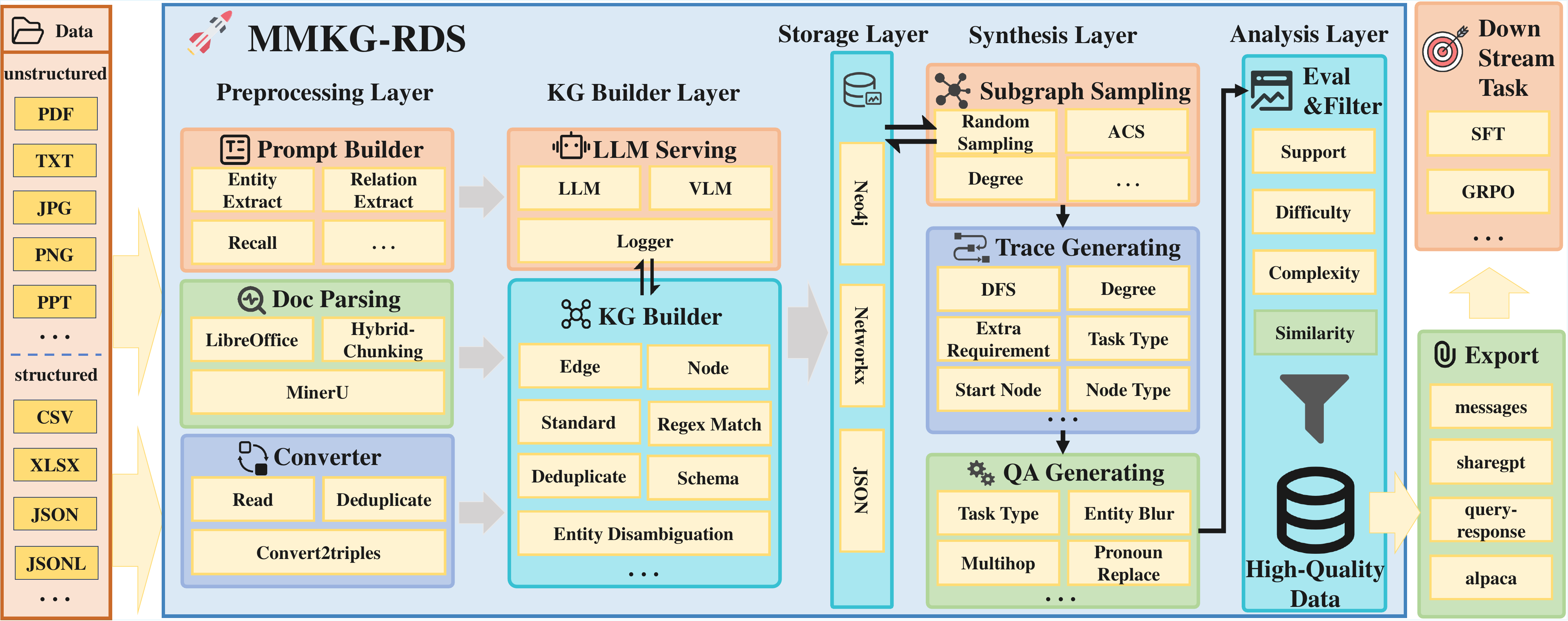}
    \caption{High-level architecture of MMKG-RDS.The system consists of five layers: Preprocessing, KG Builder, Storage, Synthesis, and Analysis. MMKG-RDS can dynamically adjust various components and task configurations based on user input, enabling the generation of high-quality, task-specific data for downstream LLM applications.}
    \label{fig:framwork}
\end{figure*}

\section{Framework Design}
The MMKG-RDS framework consists of five core components, as shown in Figure~\ref{fig:framwork}.

\textbf{Data Preprocessing Layer}: This layer integrates structured data (JSON/CSV) with unstructured documents (PDF/PNG/PPT/DOC), converting existing structured data into triplets and extracting text, images, tables, and formula information from unstructured documents.

\textbf{KG Construction Layer}: This layer supports the automated construction of knowledge graphs with customizable schemas, including constraints for entity and relationship types, as well as their attributes. It integrates the full lifecycle of entity-relation extraction, disambiguation, and normalization.

\textbf{Storage Layer}: This layer is compatible with various storage formats, including Neo4j, NetworkX, and JSON, and enables connectivity to Neo4j's visualization and analysis capabilities.

\textbf{Data Synthesis Layer}: Based on knowledge graphs, this layer synthesizes structured data through subgraph sampling, path generation, and entity fuzzification for reasoning QA generation, ensuring structural integrity, balanced distribution, and controlled task difficulty.

\textbf{Analysis Layer}: This layer provides multidimensional quality assessments, including support, difficulty, and complexity metrics, as well as data analysis features such as token distribution, task types, and domain coverage.

\begin{figure*}[t]
    \centering    \includegraphics[width=1\linewidth]{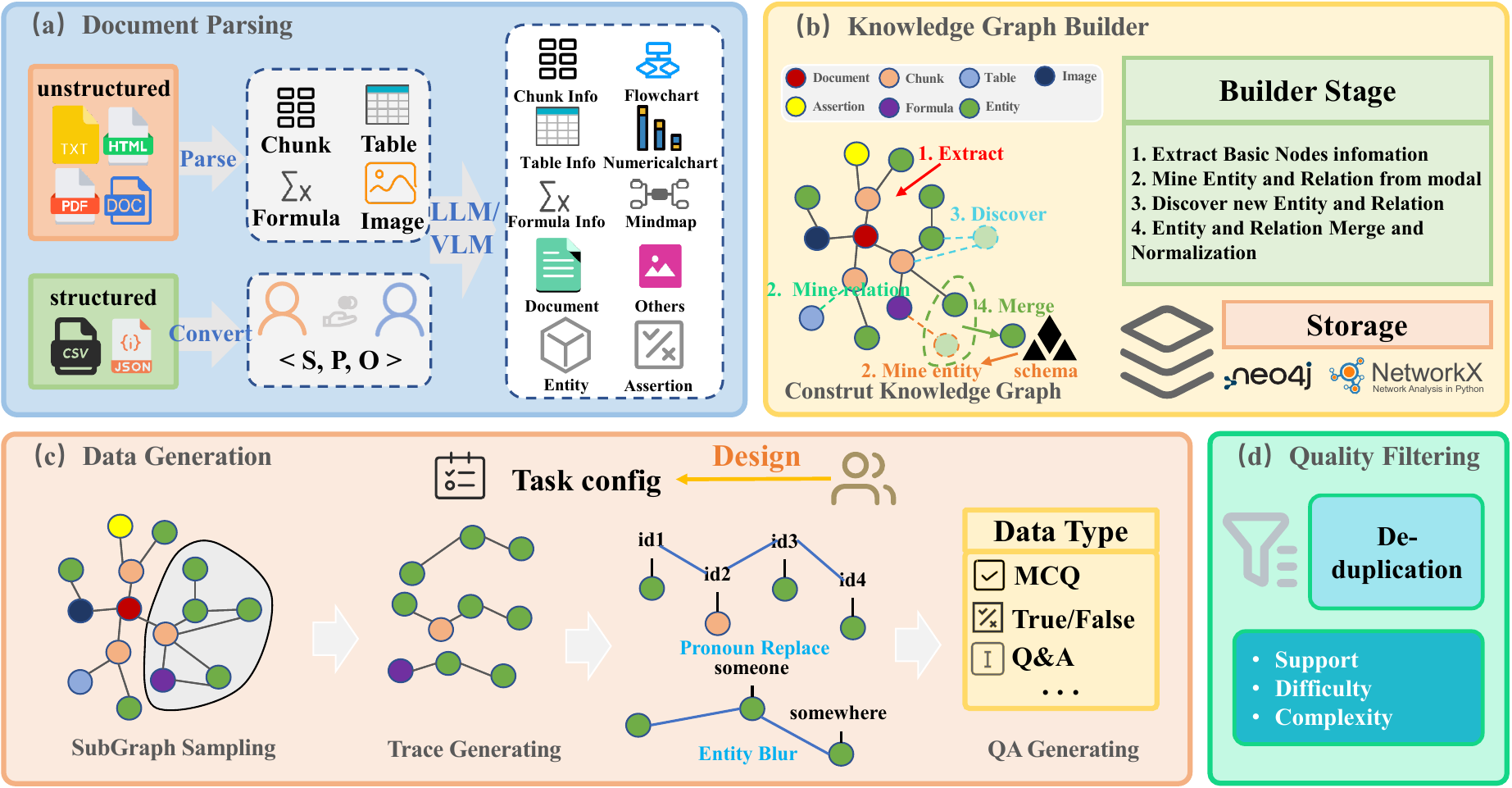}
    \caption{ MMKG-RDS comprises: (a) Document parsing: extracting and structuring multimodal documents; (b) Knowledge graph building: a five-stage construction process; (c) Data generation: synthesizing data via customizable sampling; (d) Quality filtering: deduplication, quality control, and scoring for data distribution tuning.}
    \label{fig:pipeline}
\end{figure*}

\section{Methodology}
As shown in Figure~\ref{fig:pipeline}, the proposed Multimodal Knowledge Graph-Driven Data Synthesis Framework (MMKG-RDS) can be formalized as a four-stage cascade process \( \text{MMKG-RDS} = \langle S_1, S_2, S_3, S_4 \rangle \), where the input-output and core objectives for each stage are defined as follows:

1) \textbf{Document Parsing} (\( S_1 \)): The input is a collection of heterogeneous documents \( D = \{d_1, d_2, ..., d_n\} \) (including pure text, Docx, PDF, etc.), and the output is the structured parsed result \( P = \{p_1, p_2, ..., p_n\} \) ($p_i$ being the markdown representation and intermediate results of the $i$-th element in the parsed document). The core objective is to achieve a unified representation of multi-format documents.
   
2) \textbf{Knowledge Graph Builder} (\( S_2 \)): The input is the parsed result \( P \), and the output is a multimodal knowledge graph \( \mathcal{G} = (\mathcal{V}, \mathcal{E}) \) (where \( \mathcal{V} \) is the set of nodes, and \( \mathcal{E} \) is the set of relations). In this graph, \( |\mathcal{V}_{\text{type}}| = 7 \) and \( |\mathcal{E}_{\text{type}}| = 16 \).
   
3) \textbf{Data Generation} (\( S_3 \)): The input is the knowledge graph \( \mathcal{G} \) and the user-defined sampling strategy \( \Phi \), and the output is the candidate data set \( \mathcal{D}_{\text{cand}} \). The core objective is to ensure the diversity and complexity of the data through a generation strategy.
   
4) \textbf{QA Generation} (\( S_4 \)): The input is the candidate data \( \mathcal{D}_{\text{cand}} \), and after deduplication and three-dimensional quality evaluation, the output is the high-quality data \( \mathcal{D}_{\text{final}} \).

\subsection{Document Parsing}
The framework supports parsing of both unstructured and structured documents in mainstream formats. For triplet files in JSON/CSV format \(D_{\text{struc}}\), they are converted into a triplet list. For pure text data \(D_{\text{txt}}\), users can split the text using custom separators and chunk lengths. For Docx and other rich-text documents containing text and images \(D_{\text{rich}}\), LibreOffice is used to convert them to PDFs, which are then parsed using MinerU~\cite{2409.18839}. The final representation is normalized to the format after parsing by MinerU.

\subsection{Knowledge Graph Builder}
The core goal of the Knowledge Graph Builder stage \( S_2 \) is to further mine fine-grained information from the parsed document results \( S_1(D) \) and generate a multimodal knowledge graph \( \mathcal{G} = (\mathcal{V}, \mathcal{E}) \). To construct a rich multimodal knowledge graph, we designed 7 types of nodes, as detailed in Table~\ref{tab:node_definition}. Specifically, we designed 4 stages for the construction process:

\begin{table}[htbp]
  \centering
  \resizebox{0.5\textwidth}{!}{
  \begin{tabular}{l cc}
    \toprule
    Node Type & Symbol & Fileds\\
    \midrule
    Document & $\mathcal{V}_1$ & <id, content, title, path, schema> \\
    Chunk & $\mathcal{V}_2$ & <id, content, doc\_id> \\
    Entity & $\mathcal{V}_3$ & <id, name, desc, attr, src\_id\_list> \\
    Assertion & $\mathcal{V}_4$ & <id, head, relation, tail, desc, src\_id\_list> \\
    Image & $\mathcal{V}_5$ & <id, content, caption, path, desc> \\
    Table & $\mathcal{V}_6$ & <id, content, caption, path, desc> \\
    Formula & $\mathcal{V}_7$ & <id, content, caption, desc> \\
    \bottomrule
  \end{tabular}
  }
  \caption{7 types of node definitions and fileds.}
  \label{tab:node_definition}
\end{table}

\paragraph{\textbf{stage-1 Extract Basic Nodes Information:}} For the 6 types of nodes in the design excluding \( \mathcal{V}_3 \) (Entity), we use different processing strategies to extract relevant information. The node information \( \mathcal{V} \) is listed in Table~\ref{tab:node_definition}.

1) \( \mathcal{V}_1 \) (\textbf{Document}): Retain document metadata \( (id_{d_i}, content_{d_i}, title_{d_i}, schema_{d_i}, path_{d_i}) \), where \( schema_{d_i} \) is the schema used, and \( path_{d_i} \) is the file path.
2) \( \mathcal{V}_2 \) (\textbf{Chunk}): Perform semantic-preserving merging \( c_i \oplus c_j \) for chunks that span across pages/columns in PDFs to ensure semantic integrity.
3) \( \mathcal{V}_4 \) (\textbf{Assertion}): Extract assertions in sync with entities and construct triplets \( \langle S, P, O \rangle \). Also, store the complete assertion description to resolve relationship ambiguities.
4) \( \mathcal{V}_5 \) (\textbf{Image}): Extract image elements from the document, use heuristic rules to reference chunks, and apply VLM for various categories of images such as flowcharts, numeric diagrams, and mind maps. Other image types are extracted, including captions, descriptions, and their corresponding formalized representations.
5) \( \mathcal{V}_6 \) (\textbf{Table}): Trace referenced chunks and combine HTML content parsed by MinerU as context to extract table titles, descriptions, and HTML representations.
6) \( \mathcal{V}_7 \) (\textbf{Formula}): Supplement with \( k \)-neighborhood blocks before and after the formula and trace the blocks as context to generate LaTeX representations and descriptions of formulas.
This results in the graph \( \mathcal{G}_{stage1} \).

\paragraph{\textbf{stage-2 Mine Entity and Relation from Modal:}} 
 \( \mathcal{V}_3 \)\textbf{(Entity}) are extracted from \( \mathcal{V}_2 \) (Chunk),  using large models based on the user-defined schema. To further improve the density of the multimodal knowledge graph, we supplement the extraction of entities and relations by using non-textual node content (Image, Table, Formula) and their associated information as context. The extraction approach is the same as that for text-based entities. Since these entities and relations originate from text generated by LLM/VLM based on elements, we label them as 'modal' to prevent hallucinations and errors from expanding during secondary extraction. These entities and relations are later filtered by frequency in Stage-4. The graph is updated to \( \mathcal{G}_{stage2} \).

\paragraph{\textbf{stage-3 Discover New Entity and Relation:}} Considering that native large models have low recall rates for entity-relation extraction and struggle with identifying rare entities in domain-specific data (i.e., the long-tail problem of entity distribution), we introduce a mechanism to extract new relations linked to existing entities and their contexts, thus discovering new entities. To ensure the reliability of newly discovered entities, we then attempt to recall meaningful entities based on relationships, while entities with no relevant information are excluded. The resulting graph is \( \mathcal{G}_{stage3} \).

\paragraph{\textbf{stage-4 Entity and Relation Merge and Normalization:}} Entities with different aliases should be merged. We calculate embedding similarity based on descriptive information and perform clustering, calling large models for more fine-grained grouping. Similarly, relations are processed in the same manner. Specifically, for entities that only appear once and have the 'modal' label, as well as the relations associated with them, we remove them to ensure the accuracy of the graph. Finally, the finely grouped entities and relations are merged and normalized, resulting in the final knowledge graph \( \mathcal{G} \).

\subsection{Data Generation}
To generate complex and diverse reasoning data while fully considering the significant differences between tasks, we propose a task-configured sampling synthesis strategy. The core idea is to perform flexible sampling based on graph statistics and node attribute requirements. This consists of three stages:
1) \textbf{Subgraph Sampling}: Based on a user-defined or preset sampling strategy, extract the target subgraph \( \mathcal{G'} \) from the knowledge graph \( \mathcal{G} \).
2) \textbf{Trace Generation}: Based on the sampled subgraph, combine user-defined task parameters and 32 preset path generation strategies to generate target paths, with support for custom strategy configuration. Finally, the output is formatted.
3) \textbf{QA Generation}: For different task templates, use prompt engineering, entity fuzzification, and coreference replacement to generate reasoning data that meets the requirements.

\paragraph{\textbf{Subgraph Sampling:}}
As the basic unit of synthesized data, subgraphs are extracted using the ACS backbone-enhanced chain sampling method~\citep{2508.14880}, as shown in Algorithm~\ref{alg:augmented_chain}. Nodes are selected based on in-degree, out-degree, type, and attributes.

\paragraph{\textbf{Trace Generation:}} 
The DFS algorithm is used to sample paths that meet constraint conditions from the subgraph. The system supports various configurable strategies: the initial node is selected based on the graph node's statistical information, type, and attributes. During path expansion, each time a neighboring node that meets user constraints (such as degree, type, attributes, etc.) is chosen. The sampling process terminates when the maximum node count, maximum depth, or no valid neighbors are reached.

\paragraph{\textbf{QA Generation:}}

For QA generation, we construct basic templates for open-ended, single-choice, and true/false questions (see Appendix~\ref{app:question_prompt}). To increase data difficulty, entity nodes are fuzzified by type (e.g., replacing specific names with "someone"). We also introduce a novel strategy—pronoun substitution—where entities in relationships are represented by pronouns (e.g., id), creating barriers that encourage the model to focus on key information and follow the sampled reasoning paths, thereby activating its information localization and reasoning abilities.

\subsection{Quality Filtering}
In the quality filtering phase, we first deduplicate the generated data based on question semantic similarity. We then introduce three key metrics for further data scoring: support, difficulty, and complexity. The specifics of these metrics are as follows:
\begin{itemize}
\item {\bfseries support:} 
This metric evaluates whether the reasoning process is based on the sampled path and effectively supports answer generation. Specifically, a majority voting mechanism and prompt engineering are used to deploy three independent models to judge the rationality of the reasoning path and its support for the answer. Data is accepted when at least two models confirm support.
\item {\bfseries difficulty:} Referencing s1~\cite{2501.19393}, this metric divides the data difficulty into three levels by comparing the performance difference between strong models (e.g., Qwen3-235B-A22B) and weak models (e.g., Qwen3-1.7B) on the same question. Specific standards are provided in Table~\ref{tab:hard_judge}.

\begin{table}[htbp]
\centering
\small
\setlength{\tabcolsep}{2pt}
\begin{tabular}{l *{3}{c}}
\toprule
& \textbf{simple} & \textbf{medium} & \textbf{hard} \\ \midrule

\textbf{weak\_model} & \cmark & \xmark & - \\
\textbf{strong\_model} & \cmark & \cmark & \xmark \\
\bottomrule
\end{tabular}
\caption{Definition of difficulty levels, where - denotes \cmark or \xmark.}

\label{tab:hard_judge}
\end{table}
\item {\bfseries complexity:} This metric combines various dimensions such as sentence structure complexity, task hops, required knowledge, and question length through prompt engineering to quantify the overall complexity of the data.
\end{itemize}

\begin{figure*}[htbp]
    \centering
    \includegraphics[width=\linewidth]{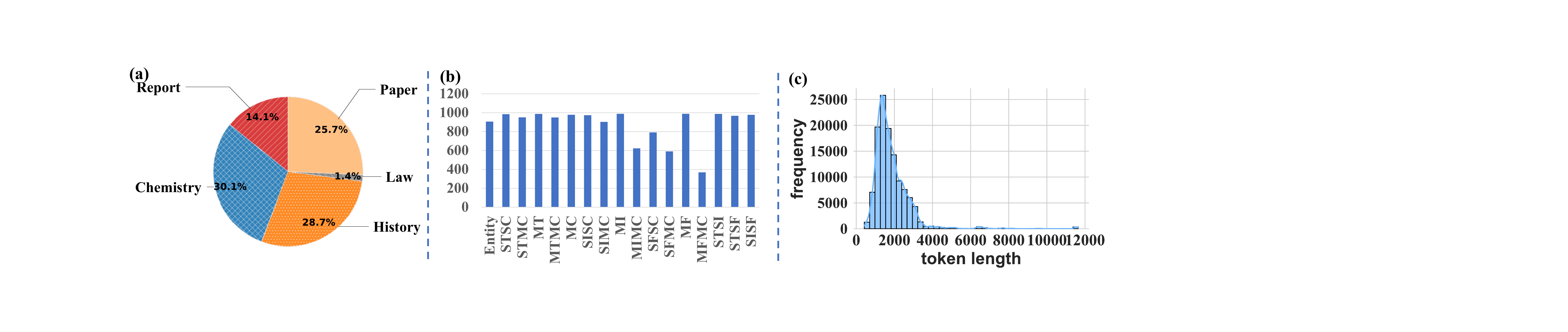}
    \caption{Statistical Information of MMKG-RDS-Bench: (a) Distribution of MMKG-RDS-Bench Across Different Domains; (b) Distribution of MMKG-RDS-Bench Across Different Tasks; (c) Distribution of Question Lengths in MMKG-RDS-Bench.}
    \label{fig:stat}
\end{figure*}

\begin{table*}[t]
\centering
\small
\setlength{\tabcolsep}{3pt}
\begin{tabular}{c l *{10}{C{0.9cm}}}
\toprule
\multirow{2}{*}{} & \multirow{2}{*}{Model} & \multicolumn{10}{c}{Task} \\ 
\cmidrule(l){3-12} 
 & & Entity & SISC & STSC & SFSC & SFMC & MC & MI & MIMC & MT & MTMC \\ 
\midrule

\multirow{8}{*}{LLM} 
& \textbf{Qwen3-0.6B} & 79.6  & 65.6  & 36.6  & 23.8  & 30.5  & 76.5  & 48.8  & 52.1  & 10.8  & 26.3  \\
& \textbf{Qwen3-1.7B} & 88.3  & 71.6  & 41.2  & 29.8  & 43.7  & 85.8  & 57.9  & 52.9  & 13.0  & 34.4  \\
& \textbf{Qwen3-4B} & 93.8  & 81.0  & 53.8  & 43.9  & 55.9  & 90.4  & 66.9  & 63.6  & 20.0  & 42.3  \\
& \textbf{Qwen3-8B} & 93.8  & 79.5  & 53.9  & 44.2  & 60.8  & 91.2  & 68.6  & 60.1  & 21.8  & 44.9  \\
& \textbf{Qwen3-14B} & 92.9  & 80.7  & \textbf{56.5} & 47.7  & 62.4  & 88.9  & \textbf{72.9} & 66.9  & \textbf{25.1} & \textbf{49.9} \\
& \textbf{Qwen3-32B} & 95.6  & 82.5  & 57.5  & 47.7  & \textbf{64.9} & \textbf{92.0} & 70.5  & \textbf{67.1} & 24.9  & 48.8  \\
& \textbf{Qwen3-30B-A3B} & \textbf{95.8} & \textbf{83.2} & 54.1  & 42.0  & 58.5  & 89.8  & 69.1  & 66.8  & 22.2  & 46.3  \\
& \textbf{Qwen3-235B-A22B} & 95.1  & 81.6  & 56.1  & \textbf{48.3} & 64.4  & 91.4  & 72.7  & 66.6  & 22.5  & 48.0  \\

\midrule
\multirow{6}{*}{VLM}
& \textbf{Qwen3-VL-2B} & 87.7  & 76.3  & 45.7  & 37.3  & 51.7  & 86.9  & 61.6  & 56.0  & 15.2  & 34.1  \\
& \textbf{Qwen3-VL-4B} & 95.5  & 82.3  & 54.1  & 43.4  & 58.6  & 92.1  & 67.9  & 65.1  & 20.8  & 46.1  \\
& \textbf{Qwen3-VL-8B} & 95.1  & 82.7  & 54.5  & 45.5  & 62.0  & \textbf{92.5} & 68.5  & 64.0  & 21.9  & 45.6  \\
& \textbf{Qwen3-VL-32B} & \textbf{96.1} & 83.0  & 55.2  & 47.3  & 65.4  & 92.4  & \textbf{74.5} & \textbf{67.5} & 23.8  & 48.4  \\
& \textbf{Qwen3-VL-30B-A3B} & 96.0  & 82.1  & 51.8  & 45.0  & 62.0  & 91.2  & 69.6  & 65.9  & 22.7  & 46.8  \\
& \textbf{Qwen3-VL-235B-A22B} & 95.8  & \textbf{84.0} & \textbf{57.7} & \textbf{50.8} & \textbf{65.8} & 91.8  & 71.8  & 66.6  & \textbf{24.8} & \textbf{51.3} \\
\bottomrule
\end{tabular}
\caption{Accuracy(\%) on different tasks across the Qwen3 full-series models. All results are averages within each task category with reasoning mode disabled.}
\vspace{-3mm}
\label{tab:tab1}
\end{table*}

\section{Experiment}

\subsection{Experimental Setup}
\textbf{Datasets.} 
We selected data from five domains: history, organic chemistry, law, stock research reports, and papers (1.8\,k pages of PDFs and 10\,k legal judgment documents). Knowledge graphs were built for each dataset using the framework, with entity and relation extraction performed by Qwen3-235B-A22B on 8 A800 GPUs. Image data were processed using Qwen3-VL-2B. All graphs were constructed in approximately 8 hours. A total of 17 task types were preset, with 500 paths sampled per task. After filtering, around 16\,k paths were obtained. QA data were generated using GPT5.1, and semantic similarity was calculated by Qwen3-Embedding-8B for deduplication. Qwen3-235B-A22B was used to calculate the support metric, resulting in 14,950 data points. Data distribution statistics are shown in Figure~\ref{fig:stat}. In (a), the relatively small proportion of questions in the legal domain is due to the fact that the data is pure text and lacks multimodal data, covering fewer task types. (b) shows the distribution of different task types, and (c) shows the frequency distribution of token lengths. The statistical information of the constructed knowledge graphs can be found in Appendix~\ref{app:graph_stat}.

\textbf{Experimental Design.} 
We designed the following three experiments to validate the effectiveness of the data generated by our framework:
1) Performance of Various Models Across Different Tasks: Validate whether different-sized models satisfy the scaling law on our synthetic data and explore the performance differences across various task types.
2) Performance of VLM Models Across Different Inputs: Validate the performance differences of VLM models with different inputs (only images, only text extracted from images, and a combination of images and text extracted from images).
3) Fine-tuning Experiments with Synthetic Data: Validate the effectiveness of fine-tuning models on synthetic data and the gain from difficulty control.

\subsection{Performance of Various Models Across Different Tasks}

\begin{figure*}
    \centering
    \includegraphics[width=1\linewidth]{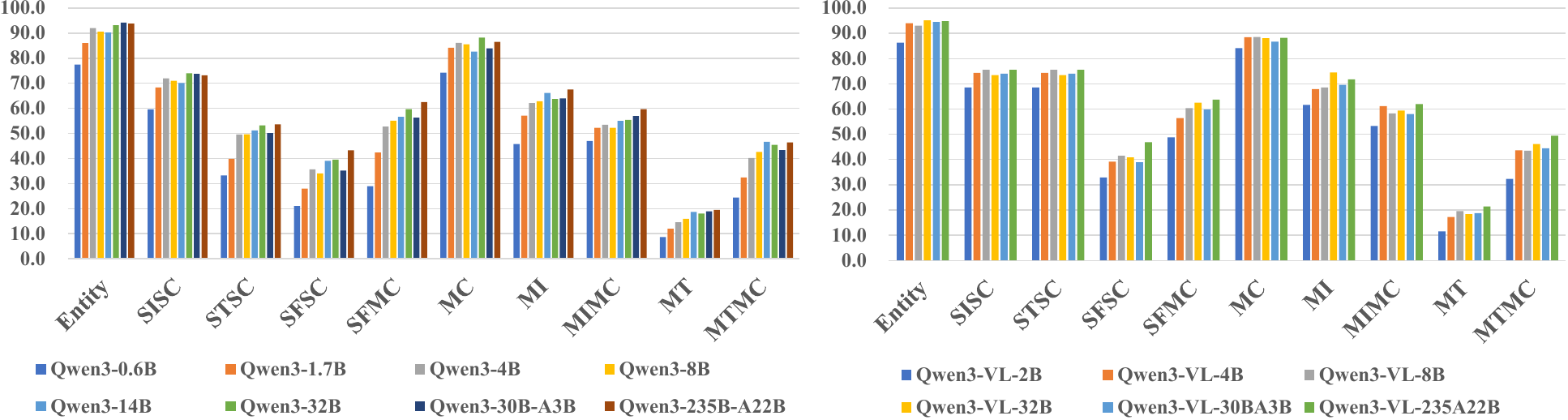}
    \caption{(left) Accuracy(\%) on different tasks across the Qwen3-series models. (right) Accuracy(\%) on different tasks across the Qwen3-VL-series models.}
    \label{fig:LLM_VLM_task}
\end{figure*}

As shown in Table~\ref{tab:tab1}, we selected 10 representative tasks from the synthetic dataset for demonstration. The task type "Entity" represents multi-hop paths that only contain entities, while the other types are task type abbreviations, where S=Single, M=Multi, I=Image, C=Chunk, F=Formula, T=Table. For example, SIMC represents multi-hop paths that contain an Image and multiple Chunks in addition to entities.

The experimental results show that the Qwen3-series models follow the scaling law for large model parameters on the synthetic data, meaning performance improves with an increase in model size, which reflects the generality of the data. Furthermore, as shown in Figure~\ref{fig:LLM_VLM_task}, there are significant performance differences across tasks: Entity and MC (multi-choice) tasks performed excellently, indicating the strong capability of LLMs in text semantic understanding and information induction. However, tasks involving images, tables, and formulas performed worse. Specifically, table tasks were the most difficult, showing that current LLMs have limited ability to process complex structures, which also involve cross-modal abilities, and indicating that the data has distinguishing features. This finding can serve as an important guide when selecting tasks for constructing complex test sets. Interestingly, for tasks containing image descriptions (MI/MIMC), the introduction of text chunks actually decreased model performance. This could be due to: 1) Increased question length leading to overall increased difficulty; 2) The image descriptions generated by Qwen3-VL-2B, which differ significantly from real text chunks and may contain some hallucinated information. Therefore, in the MIMC task, the information gain from the added text chunks was not enough to offset the noise interference.

When comparing LLM and VLM, it is clear that VLM performs better on structured tasks like tables, likely due to its stronger two-dimensional spatial perception ability, which aids in understanding chart layouts and relationships.

\subsection{Performance of VLM Models Across Different Inputs}

\begin{table}[t]
\centering
\small
\setlength{\tabcolsep}{2pt}
\begin{tabular}{l *{3}{C{1.1cm}}}
\toprule
\multirow{2}{*}{\textbf{Model}} &
\multicolumn{3}{c}{Input} \\ \cmidrule(l){2-4} 
& text & img & img+text \\ \midrule




\textbf{Qwen3-VL-2B} & \textbf{51.5}  & 49.0  & 51.1  \\
\textbf{Qwen3-VL-4B} & \textbf{60.4}  & 56.8  & 60.3  \\
\textbf{Qwen3-VL-8B} & \textbf{61.3}  & 58.4  & 60.8  \\
\textbf{Qwen3-VL-32B} & \textbf{64.0}  & 61.4  & 63.8  \\
\textbf{Qwen3-VL-30B-A3B} & \textbf{61.5}  & 57.7  & 61.1  \\
\textbf{Qwen3-VL-235B-A22B} & \textbf{65.0}  & 62.4  & 64.7 \\
\bottomrule
\end{tabular}
\caption{Accuracy(\%) on different input across the Qwen3-VL full-series models. With reasoning mode disabled, all results are averages for tasks that involve image inputs within the synthetic dataset.}
\vspace{-3mm}
\label{tab:tab2}
\end{table}

As shown in Table~\ref{tab:tab2}, we evaluated the performance of VLM models on the evaluation data under different inputs (text: text extracted only from images, img: only images, img+text: images and text extracted from images). The experimental results mostly follow the scaling law, with the performance being text > img+text > img. Introducing images led to a slight decrease in performance, which may be due to: 1) The significant differences in how different VLM models understand images without context, while the synthetic data generation relied on text extracted from images by Qwen3-VL-2B; 2) The image input increases the sequence length, and the information noise introduced may exceed the benefit of the useful information.

\subsection{Fine-tuning Experiments with Synthetic Data}

\begin{table}[t]
\centering
\small
\setlength{\tabcolsep}{2pt}
\begin{tabular}{l *{3}{C{1.6cm}}}
\toprule
\multirow{2}{*}{\textbf{Model}} &
\multicolumn{3}{c}{Mode} \\ \cmidrule(l){2-4} 
& no\_think & think & SFT(think) \\ \midrule

\textbf{Qwen3-0.6B} & 36.4 & 39.7 & \textbf{51.5}(+11.8) \\
\textbf{Qwen3-8B} & 53.4 & 59.0 & \textbf{65.6}(+6.6) \\
\textbf{Qwen3-32B} & 55.2 & 58.7 & \textbf{67.9}(+9.2) \\
\bottomrule
\end{tabular}
\caption{Performance comparison of Qwen3 models under different modes (no\_think, think, and SFT) on the synthesized dataset. Results are reported in accuracy scores.}
\vspace{-3mm}
\label{tab:tab3}
\end{table}

As shown in Table~\ref{tab:tab3}, we evenly sampled from various domains and tasks and randomly split the data into training and test sets in a 3:7 ratio. We then fine-tuned the Qwen3-series models (0.6B/8B/32B) with full parameter training. The results show that the model performance significantly improved after fine-tuning, confirming the effectiveness of the synthetic data. Before fine-tuning, the 32B model in the "think" mode performed slightly worse than the 8B model. However, after fine-tuning, the model performance significantly improved, and the 32B model outperformed the 8B model by 2.3 percentage points, demonstrating that synthetic data can effectively activate the model's reasoning ability.

To validate the effectiveness of the difficulty scoring metric, we separately extracted 1k data points each from the simple+medium and hard difficulty categories and performed fine-tuning on the Qwen3-series models (0.6B/8B/32B). Table~\ref{tab:diff_sft} shows that the accuracy improved by 2.7 percentage points after training on hard data compared to simple+medium, proving the effectiveness of the proposed difficulty metric. This also validates the effectiveness of using difficulty to filter data as shown in s1~\cite{2501.19393}.

\begin{table}[t]
\centering
\small
\setlength{\tabcolsep}{2pt}
\begin{tabular}{l *{2}{C{2.0cm}}}
\toprule
\multirow{2}{*}{\textbf{Model}} &
\multicolumn{2}{c}{Difficulty} \\ 
\cmidrule(l){2-3} 
& simple+medium & hard \\ \midrule

\textbf{Qwen3-0.6B} & 52.0 & \textbf{54.7}(+2.7) \\
\textbf{Qwen3-8B} & 67.8 & \textbf{70.5}(+2.7) \\
\textbf{Qwen3-32B} & 69.4 & \textbf{72.7}(+3.3) \\
\bottomrule
\end{tabular}
\caption{Accuracy(\%) after SFT on 1\,k samples selected from different difficulty in the training set.}
\vspace{-3mm}
\label{tab:diff_sft}
\end{table}

\section{Conclusion}
This paper proposes the MMKG-RDS framework, a flexible system for generating synthetic reasoning data by leveraging multimodal knowledge graphs. It supports fine-grained knowledge extraction, customizable path sampling, and multidimensional data quality scoring. Using this framework, we constructed the MMKG-RDS-Bench, a multi-domain reasoning dataset covering 5 domains and 17 task types (14,950 instances). Evaluation results show that existing models still struggle with cross-modal complex reasoning tasks like tables and formulas, indicating that this framework can create challenging evaluation benchmarks. Training experiments demonstrate that the generated reasoning datasets are effective for model training.

\section*{Limitation}

It is well-known that constructing a perfect framework for synthetic reasoning data generation for large models is no easy task. This work contributes a reasoning data synthesis framework from the perspective of fine-grained multimodal knowledge graphs, but it still has several limitations.

Firstly, in the aspect of knowledge graph construction, schema is a core issue. The current framework ensures the accuracy and effectiveness of extracted information through constraints based on schemas and existing entity relationships. However, schemas still need to be manually constructed, which limits the applicability of the framework. In the future, we plan to explore dynamic schema construction methods and integrate them into this framework.

Secondly, the quality of document parsing has a significant impact on the overall performance. MMKG-RDS relies on MinerU for document parsing, but MinerU still has some issues, such as splitting a complete paragraph into two parts if it is interrupted by an image, or handling cross-page tables. Additionally, for images containing multiple sub-graphs, MinerU may split the image into several sub-graphs, leaving the caption only for the first one. Although our framework addresses some of these document parsing issues, it is still not complete.

Thirdly, this work heavily relies on the performance and stability of existing VLMs and LLMs for multimodal information extraction and text information extraction. Using multimodal models to extract image information may introduce hallucination issues, leading to greater errors in extracting entities from non-text nodes. To mitigate this, we use similar information from other parts of the document to verify the consistency of extracted entities, as entities that appear in two or more places are considered more reliable. However, some entities in images may not be verifiable by other parts of the document and are filtered out by our framework. These entities might be important, but it is difficult to verify whether they are erroneous or hallucinated.

Fourthly, there is room for improvement in the system's execution efficiency. Although we have incorporated asynchronous processing and random load balancing, the graph construction process is still relatively slow. This is due to factors such as large model response times and document length, making the overall extraction process time-consuming. Thus, further optimization of prompt length and quality is needed in future work.

Finally, the breadth of the data and further evaluation on larger models could be explored in more depth.



\bibliography{acl_latex}

\appendix

\clearpage
\section*{\Large Appendix}
\label{sec:appendix}


\section{Knowledge Graph Visualization}

\begin{figure*}[htbp]
    \centering
    \includegraphics[width=\linewidth]{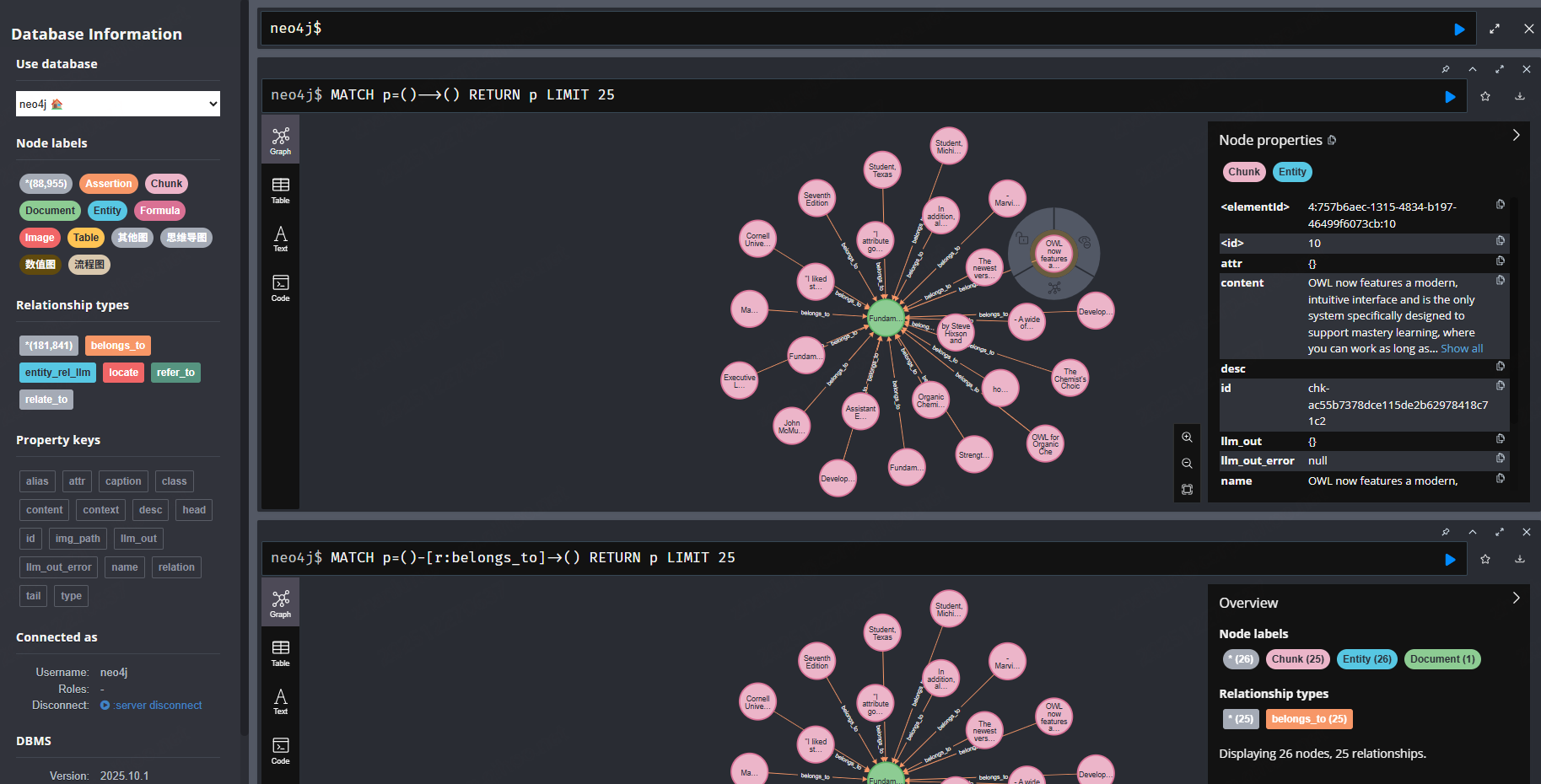}
    \caption{Neo4j-based Knowledge Graph in the Chemistry Domain}
    \label{fig:neo4j}
\end{figure*}

As shown in Figure~\ref{fig:neo4j}, we visualized the knowledge graph constructed using a basic organic chemistry textbook. It consists of a total of 88,955 nodes and 181,841 edges.

\section{Graph Statistics} \label{app:graph_stat}
As shown in Table~\ref{tab:node_domain_cnt}, we present the statistics for the number of nodes in various categories across the five selected domains. Among them, history and chemistry are professional domain books with more than 600 pages, while law is an XLSX file containing about 10\,k legal documents.

\begin{table}[htb]
\centering
\small
\setlength{\tabcolsep}{2pt}
\begin{tabular}{l *{5}{c}}
\toprule
\multirow{2}{*}{\textbf{Node}} &
\multicolumn{5}{c}{Domain} \\ \cmidrule(l){2-6} 
& law & report & paper & history & chemistry \\ \midrule

Document & 1 & 10 & 10 & 1 & 1 \\
Chunk & 9961 & 697 & 1748 & 2594 & 7847 \\
Assertion & 30091 & 12804 & 24164 & 32867 & 58703 \\
Entity & 15183 & 5097 & 10140 & 12788 & 19568 \\
Table & 0 & 98 & 104 & 7 & 43 \\
Image & 0 & 118 & 85 & 46 & 2667 \\
Formula & 0 & 0 & 59 & 0 & 126 \\
\bottomrule
\end{tabular}
\caption{Statistics on the number of knowledge graph nodes in different domains.}
\label{tab:node_domain_cnt}
\vspace{-3mm}
\end{table}

\begin{table}[htbp]
\centering
\small
\setlength{\tabcolsep}{2pt}
\begin{tabular}{l *{5}{c}}
\toprule
\multirow{2}{*}{\textbf{Edge}} &
\multicolumn{5}{c}{Domain} \\ \cmidrule(l){2-6} 
        &   law & report & history & paper & chemistry \\
\midrule
chk$\to$doc &  9961 &    697 &    2594 &  1748 &     7847 \\
chk$\to$fml &     0 &      0 &       0 &     5 &        0 \\
chk$\to$img &     0 &      0 &       0 &     1 &       22 \\
chk$\to$tbl &     0 &      5 &       0 &    85 &       54 \\
ent$\to$chk & 35589 &   4909 &   23684 & 13382 &    31231 \\
ent$\to$ent & 30091 &  12804 &   32867 & 24164 &    58703 \\
ent$\to$fml &     0 &      0 &       0 &   781 &     1459 \\
ent$\to$img &     0 &   1257 &     459 &  1053 &    20574 \\
ent$\to$tbl &     0 &   2106 &      90 &  2314 &     1059 \\
ass$\to$chk & 30091 &   6730 &   32190 & 18451 &    37117 \\
ass$\to$fml &     0 &      0 &       0 &   730 &     1470 \\
ass$\to$img &     0 &   1911 &     477 &  1411 &    18741 \\
ass$\to$tbl &     0 &   4163 &     200 &  3572 &      728 \\
fml$\to$doc &     0 &      0 &       0 &    59 &      126 \\
img$\to$doc &     0 &    118 &      46 &    85 &     2667 \\
tbl$\to$doc &     0 &     98 &       7 &   104 &       43 \\
\bottomrule

\end{tabular}
\caption{Counts of edge types between different node categories in the multimodal knowledge graph. Node category abbreviations are: doc (Document), chk (Chunk), ent (Entity), ass (Assertion), fml (Formula), img (Image), tbl (Table).}
\label{tab:edge_domain_cnt}
\vspace{-3mm}
\end{table}
As shown in Table~\ref{tab:edge_domain_cnt}, we present the statistics for the number of different types of edges in the synthesized graphs across the five domains. It is evident that edges such as \textbf{chk->fml}, \textbf{chk->img}, and \textbf{chk->tbl} are quite sparse. This is due to the lack of special annotations in most book documents, and regex-based rule matching can only identify a small number of chunks with these references, resulting in sparse edges for these nodes. Consequently, the knowledge graph is less likely to sample paths involving these elements, making it more difficult to extract fine-grained information. Therefore, we extract entities from Image, Table, and Formula to build denser connections, enabling more detailed extraction of these rich document elements.

\section{Task Information}
As shown in Table~\ref{tab:qa_task}, we list the 32 pre-defined task types within our framework.

\section{Subgraph sampling}
The sampling algorithm supported by the framework is shown in Algorithm~\ref{alg:augmented_chain}.

\begin{algorithm}[t]
\small
\caption{Augmented Chain Sampling}
\label{alg:augmented_chain}
\renewcommand{\algorithmicrequire}{\textbf{Input:}}
\renewcommand{\algorithmicensure}{\textbf{Output:}}
\newcommand{\BREAK}{\STATE \textbf{break}}

\begin{algorithmic}[1]
\REQUIRE
    Graph $\mathcal{G} = (\mathcal{V}, \mathcal{E})$, sample size $k$
\ENSURE
    Sampled subgraph $\mathcal{G'} = (\mathcal{V'}, \mathcal{E'})$
    \STATE $\text{start\_node}, \text{end\_node} \gets \text{SelectDistantNodes}(G)$
    \STATE $\text{backbone\_path} \gets \text{ExtractBackbonePath}(\mathcal{G}, \text{start\_node}, \text{end\_node})$
    \STATE $\mathcal{V'} \gets \{\,v \mid v \in \text{backbone\_path}\,\}$
    \FOR{each $v_i \in \text{backbone\_path}$}
        \STATE $\mathcal{N}_i \gets \{\,v_j \in \mathcal{V} \mid (v_i, v_j) \in \mathcal{E}\,\} \setminus \mathcal{V'}$
        \STATE $\text{budget} \gets \min(\text{RandomInt}(1, 2), |\mathcal{N}_i|, k - |\mathcal{V'}|)$
        \IF{$\text{budget} > 0$}
            \STATE $\mathcal{V'} \gets \mathcal{V'} \cup \text{RandomSample}(\mathcal{N}_i, \text{budget})$
        \ENDIF
        \IF{$|\mathcal{V'}| \geq k$}
            \BREAK
        \ENDIF
    \ENDFOR
    \STATE $\mathcal{E'} \gets \{\,(u, v) \in E \mid u \in \mathcal{V'} \land v \in \mathcal{V'}\,\}$
    \RETURN $\mathcal{G'} = (\mathcal{V'}, \mathcal{E'})$
\end{algorithmic}
\end{algorithm}

\begin{table*}[t!]
\centering
\small
\setlength{\tabcolsep}{6pt}
\begin{tabular}{ll}
\toprule
\textbf{Modality Category} & \textbf{Task Types} \\
\midrule
\multirow{5}{*}{\textbf{Table-centric QA}} 
    & Single-Table QA \\
    & Single-Table Multi-hop QA \\
    & Single-Table \& Single-Chunk (Text) QA \\
    & Single-Table \& Multi-Chunk (Text) QA \\
    & Multi-Table QA \\
    & Multi-Table Multi-hop QA\\
    & Multi-Table \& Multi-Chunk (Text) QA \\
\midrule
\multirow{4}{*}{\textbf{Text-only QA}}
    & Single-Chunk (Text) QA \\
    & Multi-Chunk (Text) QA \\
    & Single-Chunk (Text) Multi-hop QA \\
    & Multi-Chunk (Text) Multi-hop QA \\
\midrule
\multirow{5}{*}{\textbf{Formula-centric QA}}
    & Single-Formula QA \\
    & Single-Formula Multi-hop QA \\
    & Single-Formula \& Single-Chunk (Text) QA \\
    & Single-Formula \& Multi-Chunk (Text) QA \\
    & Multi-formula QA \\
    & Multi-formula Multi-hop QA \\
    & Multi-formula \& Multi-Chunk (Text) QA \\
\midrule
\multirow{5}{*}{\textbf{Image-centric QA}}
    & Single Image QA \\
    & Single Image Multi-hop QA \\
    & Single Image \& Single-Chunk (Text) QA \\
    & Single Image \& Multi-Chunk (Text) QA \\
    & Multi-Image QA \\
    & Multi-Image Multi-hop QA \\
    & Multi-Image \& Multi-Chunk (Text) QA \\
\midrule
\multirow{4}{*}{\textbf{Cross-modality QA}}
    & Single-Table \& Single Image QA \\
    & Single-Table \& Single Image Multi-hop QA \\
    & Single-Table \& Single-Formula QA \\
    & Single-Table \& Single-Formula Multi-hop QA \\
    & Single Image \& Single-Formula QA \\
    & Single Image \& Single-Formula Multi-hop QA \\
\midrule
\textbf{Entity-centric QA}
    & Pure Entity Multi-hop QA \\
\bottomrule
\end{tabular}
\caption{Taxonomy of predefined Question Answering Tasks by Modality.}
\label{tab:qa_task}
\end{table*}

\section{Case Study}
\begin{figure}[htb]
	\centering
	\includegraphics[width=\linewidth]{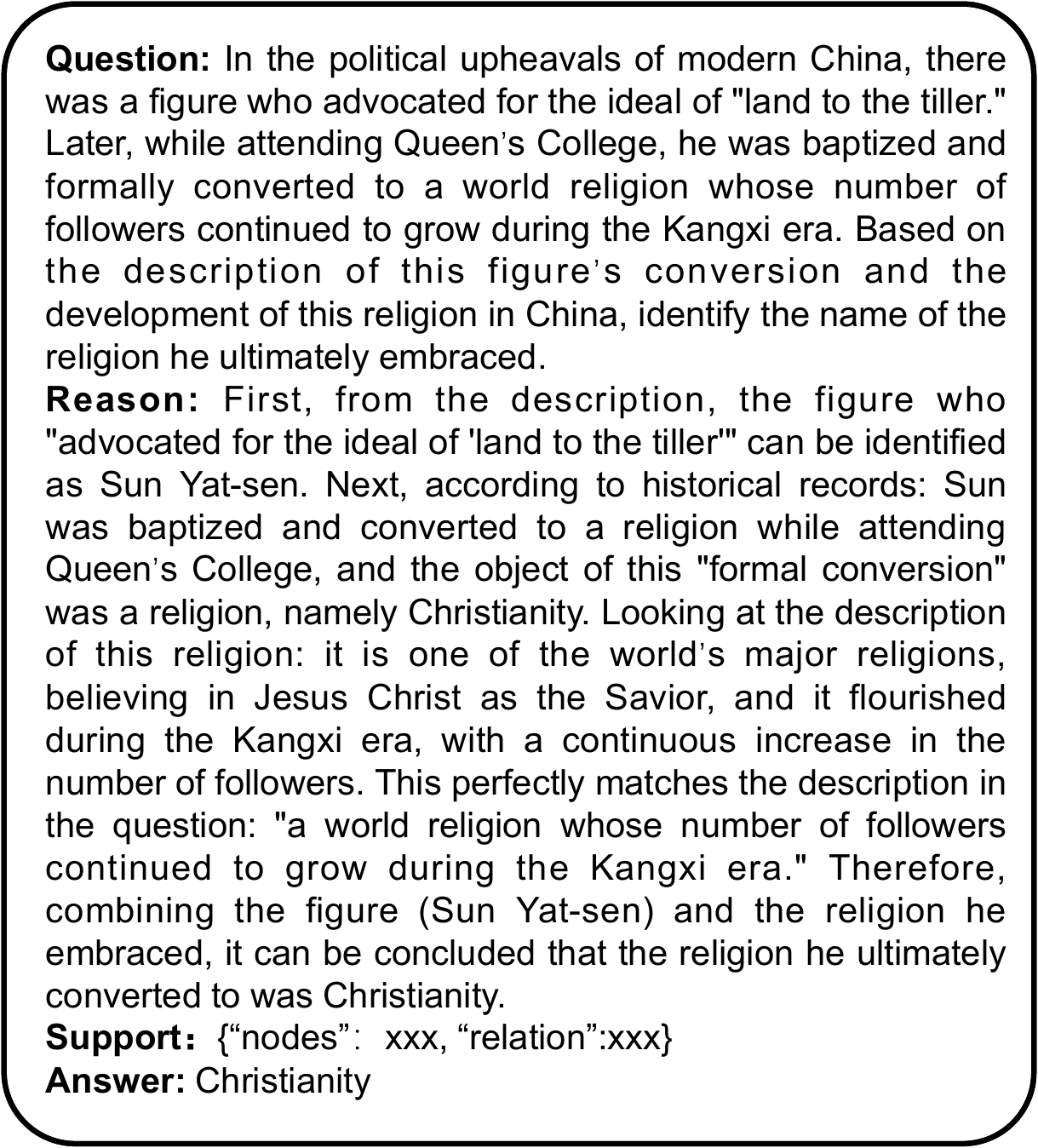}
	\caption{Case 1 from MMKG-RDS-bench}
	\label{fig:case1}
\end{figure}

\begin{figure}[htb]
	\centering
	\includegraphics[width=\linewidth]{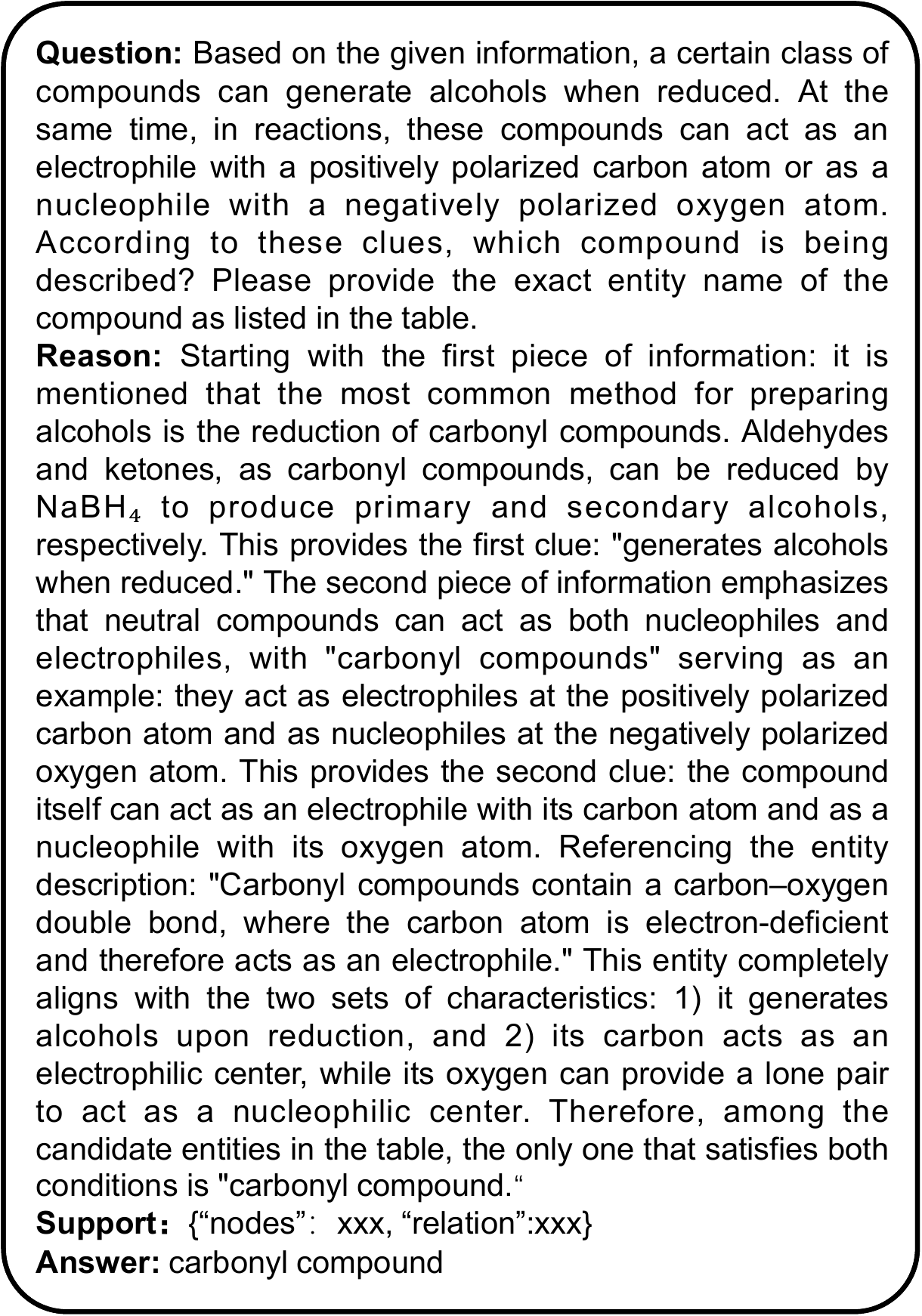}
	\caption{Case 2 from MMKG-RDS-bench}
	\label{fig:case2}
\end{figure}

Figure~\ref{fig:case1} demonstrates an example of data generated from sampling a path in the history graph. The logical steps are rigorous, and the intermediate entity implied by the chain is correctly inferred, with the answer being accurate. This somewhat proves the value of using referential word replacement. Similarly, Figure~\ref{fig:case2} shows data generated from sampling a path in the chemistry graph, where basic organic chemistry knowledge can also validate the correctness of the logic and the answer.

\begin{figure*}[htb]
	\centering
	\includegraphics[width=0.8\linewidth]{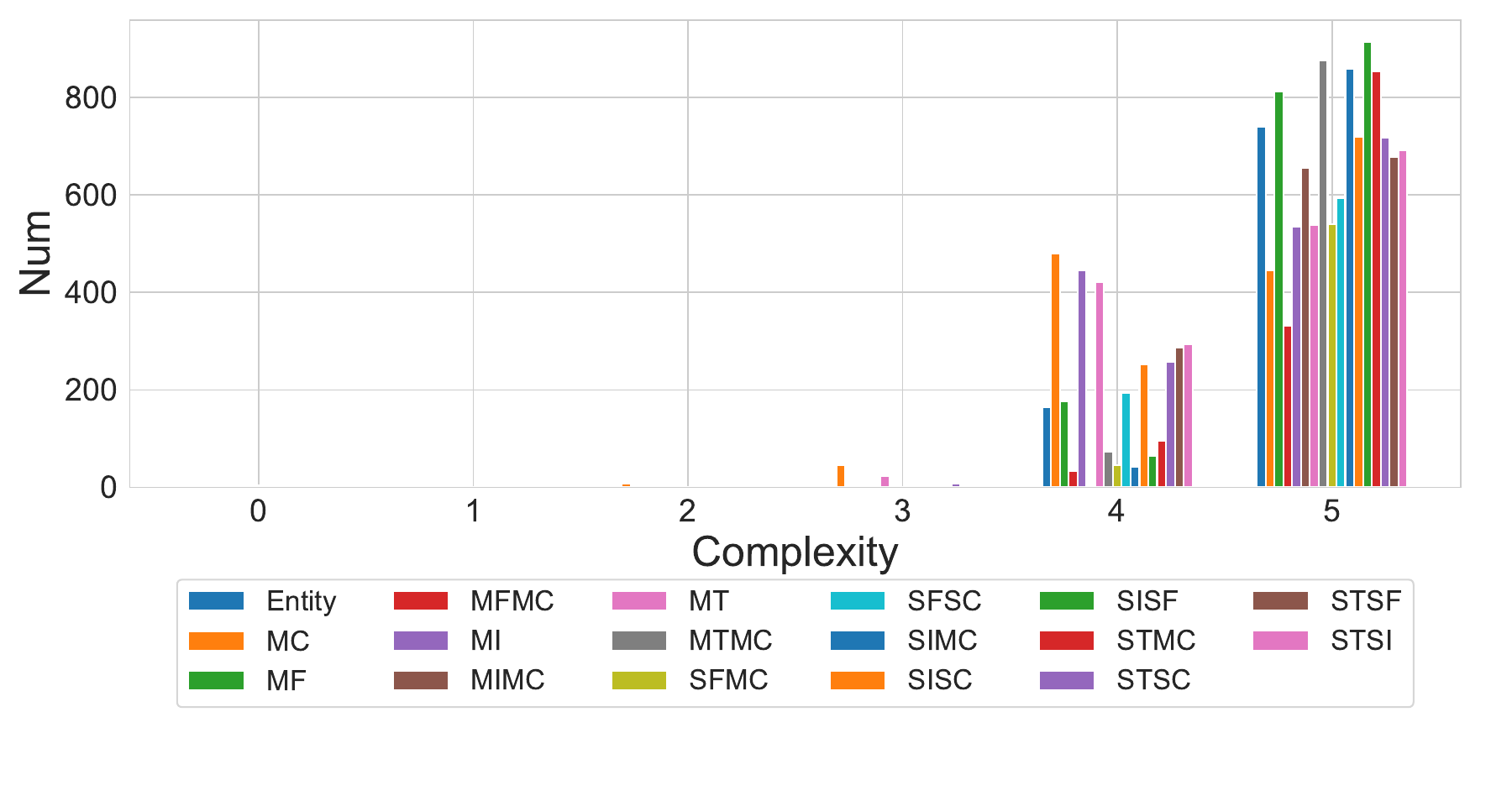}
    \caption{Distribution of Complexity on MMKG-RDS-bench.}
	\label{fig:comp_task_distr}
\end{figure*}

\begin{figure*}[htb]
	\centering
	\includegraphics[width=0.8\linewidth]{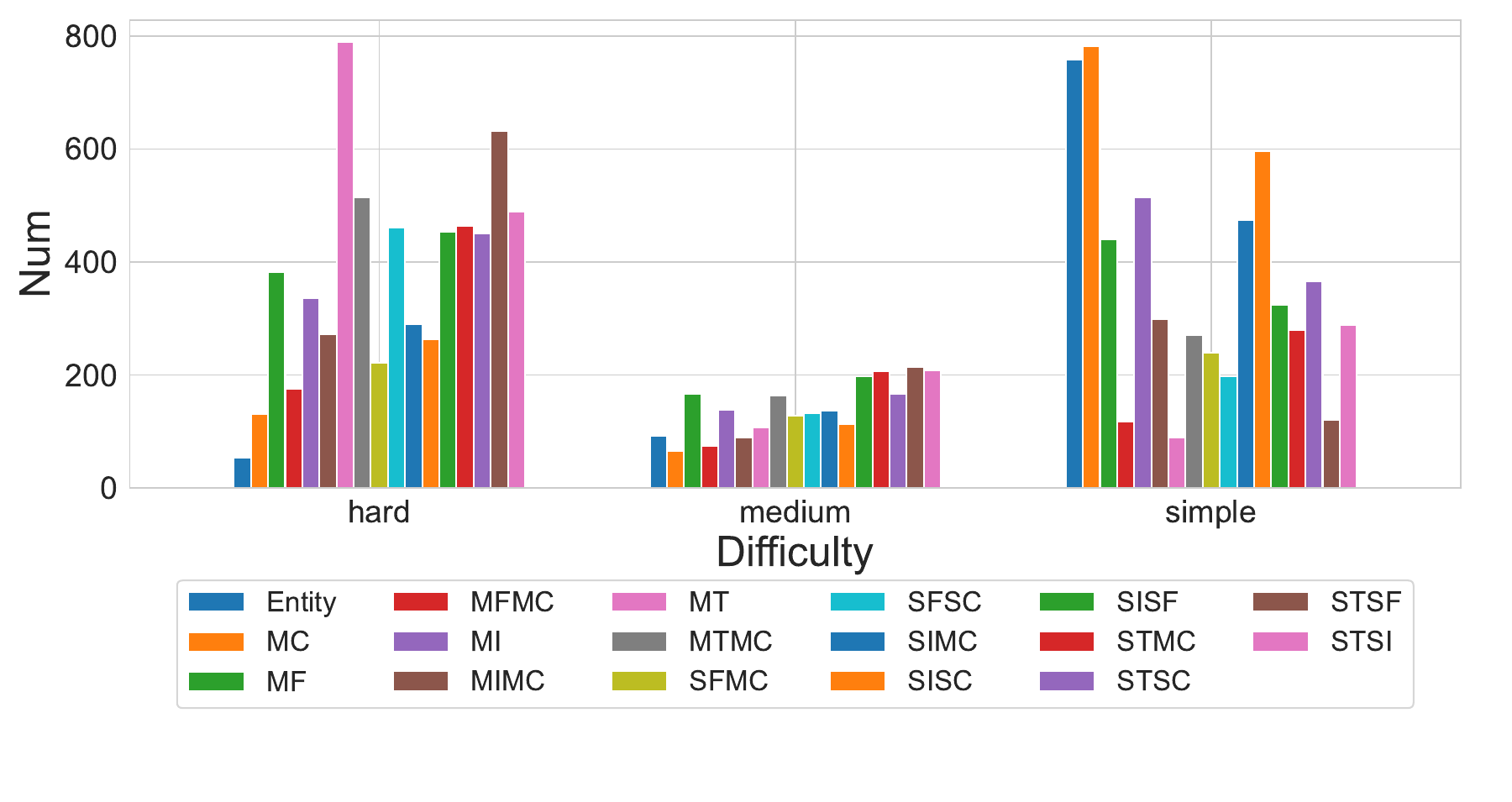}
    \caption{Distribution of Difficulty on MMKG-RDS-bench.}    
	\label{fig:diff_task_distr}
\end{figure*}

\section{Difficulty \& Complexity Analysis}
As shown in Figures~\ref{fig:comp_task_distr} and~\ref{fig:diff_task_distr}, we present the distributions of complexity and difficulty across our dataset.

In Table~\ref{tab:diff_complexs}, we show the performance of the Qwen3 series models on tasks of varying difficulty and complexity. The Qwen3-1.7B and Qwen3-235B-A22B models participated in the difficulty evaluation, and their results should not be considered in the difficulty comparison. The table analysis indicates that data at different difficulty and complexity levels generally follow the scaling law, with significant performance differentiation at each level. However, there is still an instance where the 14B model outperforms the 32B model on hard difficulty data. Upon analyzing this, we found that even though Qwen3-235B-A22B answered incorrectly on the hard difficulty problem set, it still achieved a 16.5\% correct rate during the evaluation. This may be due to the mechanism's lack of robustness when evaluating difficulty based on a single evaluation.

\begin{table*}[htbp]
\centering
\small
\setlength{\tabcolsep}{3pt}
\begin{tabular}{cl *{3}{C{0.9cm}} | *{3}{C{0.9cm}}}
\toprule
\multirow{2}{*}{} & \multirow{2}{*}{Model} &
\multicolumn{3}{c}{Difficulty} & \multicolumn{3}{c}{Complexity} \\ 
\cmidrule(l){3-8} 

 & & simple & medium & hard & 3 & 4 & 5 \\ \midrule

 \multirow{8}{*}{LLM} 


& \textbf{Qwen3-0.6B} & 73.5  & 28.9  & 13.1  & 47.3  & 41.6  & 40.2  \\
& \textcolor{blue}{\textbf{Qwen3-1.7B}} & 97.0 & 12.5  & 12.1  & 47.3  & 47.3  & 47.0  \\
& \textbf{Qwen3-4B} & 93.9  & 66.3  & 22.3  & 64.5  & 58.4  & 58.9  \\
& \textbf{Qwen3-8B} & 94.7  & 68.5  & 23.1  & 63.4  & 59.0  & 60.1  \\
& \textbf{Qwen3-14B} & 95.7  & 78.3  & 25.5  & 62.4  & 61.7  & 63.3  \\
& \textbf{Qwen3-32B} & 95.1  & 80.6  & 23.4  & 64.5  & 62.4  & 62.0  \\
& \textbf{Qwen3-30B-A3B} & 96.5  & 79.9  & 21.5  & 58.1  & 60.8  & 62.1  \\
& \textcolor{blue}{\textbf{Qwen3-235B-A22B}} & 99.2  & 97.0  & 16.5  & 66.7  & 61.8  & 64.0 \\

\midrule
\multirow{6}{*}{VLM}
& \textbf{Qwen3-VL-2B} & 91.1  & 42.5  & 16.6  & 59.1  & 52.2  & 51.2  \\
& \textbf{Qwen3-VL-4B} & 95.9  & 72.9  & 21.4  & 59.1  & 59.1  & 60.7  \\
& \textbf{Qwen3-VL-8B} & 96.1  & 77.4  & 21.7  & 63.4  & 60.6  & 61.5  \\
& \textbf{Qwen3-VL-32B} & 97.8  & 86.4  & 23.0  & 67.7  & 63.0  & 64.2  \\
& \textbf{Qwen3-VL-30B-A3B} & 96.9  & 79.6  & 20.6  & 58.1  & 60.7  & 61.8  \\
& \textbf{Qwen3-VL-235B-A22B} & 98.7  & 91.9  & 22.4  & 69.9  & 63.3  & 65.5 \\
  
\bottomrule

\end{tabular}
\caption{Accuracy(\%) on different difficulty and complexity across the Qwen3 full-series models. All results are averages within each task category with reasoning mode disabled. Among these models, \textcolor{blue}{Qwen3-1.7B} and \textcolor{blue}{Qwen3-235B-A22B} participated in the difficulty level evaluation.}
\vspace{-3mm}
\label{tab:diff_complexs}

\end{table*}

\section{Main Prompt}

\subsection{Question}\label{app:question_prompt}
The following are the prompts for the three predefined question templates within the framework.

\begin{tcolorbox}[breakable, title=MCQ, colframe=blue!75!black, colback=white, fonttitle=\bfseries, before skip=1pt, after skip=1pt, fontupper=\linespread{0.8}\selectfont]
{\footnotesize

You are a professional complex question-answering data synthesis expert, skilled at transforming single-hop and multi-hop relationships in knowledge graphs into highly concealed and highly challenging single-choice questions.

Please generate a high-quality multiple-choice question based on the provided knowledge graph subgraph or path information:

\textbf{\{graph\_context\}}

\noindent\textbf{Generation Requirements:}
\begin{enumerate}
    \item The question must be \textbf{complex and intricate}, derived by tracing the relationship chains of multiple entities, and comprehensively utilizing multiple entities and relationships in the knowledge graph.
    \item The question must be a \textbf{single-choice question (A/B/C/D)} with exactly one correct answer. For the three distractor options, avoid using other entity names on the path; instead, use relevant misleading options.
    \item The question must be \textbf{natural and fluent}, avoiding mechanical phrasing and using natural language expression.
    \item The question can be \textbf{contextually packaged} based on a specific real-world scenario, rather than explicitly exposing reasoning steps, and must never reveal the name of the answer.
    \item The reasoning chain can utilize additional attribute information of each entity on the path.
    \item The synthesized reasoning chain must be \textbf{coherent}, simulating real human reasoning processes with natural logical connectives, rather than mechanically listing entities and relationships. It should be written as a coherent thinking process.
    \item The answer must be the exact option (A, B, C, or D) corresponding to the endpoint entity of the provided path, concise, specific, and unique.
    \item The generated content must not contain redundant information such as "according to the knowledge graph", "based on a certain relationship", etc.
    \item If the provided path or subgraph contains only one entity node, generate a reasonable relevant question and reasoning process based on the entity's attributes.
    \item The final output must be in strict JSON format.
\end{enumerate}

Please generate a question-answer pair that meets the requirements and return it in JSON format:
\begin{Verbatim}[breaklines=true, showspaces=false]
{
    "question": "Generated complex question with four options A/B/C/D",
    "answer": "The only correct option, e.g., A, B, C, D",
    "reasoning_path": "CoT thinking or reasoning process for question design, which must be natural and coherent"
}
\end{Verbatim}

}
\end{tcolorbox}
\begin{tcolorbox}[breakable, title=Q\&A, colframe=blue!75!black, colback=white, fonttitle=\bfseries, before skip=1pt, after skip=1pt, fontupper=\linespread{0.8}\selectfont]
{\footnotesize

You are a {professional complex question-answering data synthesis expert}, skilled at transforming {multi-hop relationships} in knowledge graphs into natural and fluent reasoning-based questions.

Please generate a high-quality multi-hop reasoning question-answer pair based on the provided knowledge graph subgraph or path information, along with relevant carrier evidence (images/tables/formulas, etc.).

\vspace{1em}
Knowledge Graph Subgraph or Path Information:

\textbf{\{graph\_context\}}

\begin{itemize}
    \item \textbf{Entity Nodes}: Real entities in the knowledge graph, such as people, locations, models, and tasks.
    \item \textbf{Carrier Nodes} (Table, Image, Figure, etc.): Used to provide relevant context for preceding entities and cannot be directly used as answers.
    \item \textbf{Relationships}: Relationships like \texttt{belong\_to} and \texttt{locate\_at} only indicate attribution or positioning and do not participate in the logical deduction of the reasoning chain.
\end{itemize}

\vspace{1em}
\noindent\textbf{Generation Requirements}
\begin{enumerate}
    \item \textbf{Question Design Principles}:
        \begin{itemize}
            \item The question must be derived by reasoning through relationships among multiple entities, reflecting the complexity of cross-entity, multi-hop logic. Avoid consecutive multiple questions.
            \item The question must be natural and fluent, avoiding mechanical phrasing and using natural language expression.
            \item The question must be contextually packaged rather than explicitly exposing reasoning steps. It is strictly prohibited to reveal the answer name or key words contained in the answer.
            \item If charts or images contain numerical values, trends, comparative relationships, etc., ensure that the generated question includes sufficient carrier-related context. Otherwise, the model cannot perform reasoning and answering. You need to incorporate enough information into the question as context to ensure the model can only derive the correct answer by "understanding the chart".
            \item \textbf{When the path includes image or table nodes}: Design reasoning types related to the carrier content, such as numerical comparison, trend changes, etc.
            \item The question should be based on the information presented by the carrier, but the final answer should still be the exact name of the entity connected to the carrier or its explicit attribute.
        \end{itemize}
    
    \item \textbf{Reasoning Process Requirements}:
        \begin{itemize}
            \item The reasoning chain can utilize additional attribute information of each entity on the path.
            \item The synthesized reasoning chain must be coherent, simulating real human reasoning processes with natural logical connectives. Do not explicitly label triple relationships; instead, write it as a coherent thinking process rather than mechanically listing entities and relationships.
            \item The generated content must not contain redundant information such as "according to the knowledge graph", "based on a certain relationship", "according to entity information", "the complete reasoning chain is as follows", or "according to a certain image/table".
            \item \textbf{When the endpoint is a chart/table/formula}: This node and its related attributes are only carrier evidence used to derive the exact name or specific attribute value of the \textbf{entity preceding the carrier}.
            \item Relationships like \texttt{belong\_to} and \texttt{locate\_at} only indicate such attribution relationships and are not used as reasoning relationships in the actual reasoning process. Analyze how the relevant carrier content supports or defines the preceding entity.
            \item When the path includes chart/image/table nodes, their content should be used as auxiliary evidence to support or limit the reasoning direction.
        \end{itemize}
    
    \item \textbf{Answer Requirements}:
        \begin{itemize}
            \item If the path endpoint contains carrier evidence such as "image/table/formula": The answer should be the exact name of the \textbf{entity preceding the carrier} (e.g., for path A$\rightarrow$B$\rightarrow$[Chart], the answer = B or its attribute value). Avoid overly long answers.
            \item It is strictly prohibited to use the carrier name or its related attributes as the answer.
            \item The answer must be unique, specific, and avoid ambiguity or polysemy.
            \item The answer should preferably not be numerical; it can be a trend change obtained through multi-step reasoning or the name of a relevant entity in the chart.
        \end{itemize}
    
    \item \textbf{Special Handling}:
        \begin{itemize}
            \item If the path contains only a single entity node:
                \begin{itemize}
                    \item For ordinary entities: Generate a question-answer pair based on its attributes.
                \end{itemize}
        \end{itemize}
\end{enumerate}

\vspace{1em}
\noindent\textbf{Output Specifications}
\begin{itemize}
    \item Return strictly in \textbf{JSON format} with no additional text (including markers like "```json").
    \item Required fields:
\end{itemize}

\begin{Verbatim}[breaklines=true, showspaces=false]
{
    "question": "A complex natural language question",
    "answer": "A concise, specific, and unique answer",
    "reasoning_path": "The reasoning process for question design, which is a CoT thinking or reasoning process that must be natural and coherent"
}
\end{Verbatim}

}
\end{tcolorbox}

\begin{tcolorbox}[breakable, title=T\&F, colframe=blue!75!black, colback=white, fonttitle=\bfseries, before skip=1pt, after skip=1pt, fontupper=\linespread{0.8}\selectfont]
{\footnotesize

You are a professional complex question-answering data synthesis expert, skilled at transforming single-hop and multi-hop relationships in knowledge graphs into highly concealed and highly challenging true/false questions.

Please generate a high-quality true/false question based on the provided knowledge graph subgraph or path information:

\textbf{\{graph\_context\}}

\noindent\textbf{Generation Requirements:}
\begin{enumerate}
    \item The question must be \textbf{complex and intricate}, derived by tracing the relationship chains of multiple entities, and comprehensively utilizing multiple entities and relationships in the knowledge graph—avoiding simple direct assertions of single relationships.
    \item The question must be a \textbf{true/false question} with an unambiguous answer (True/False). The judgment must strictly align with the logical relationships and attribute information in the knowledge graph, without ambiguity.
    \item The question must be \textbf{natural and fluent}, avoiding mechanical phrasing (e.g., direct listing of entities and relationships) and adopting natural language expression consistent with real-world communication scenarios.
    \item The question can be \textbf{contextually packaged} based on a specific real-world scenario (e.g., academic research, daily applications, professional practices), rather than explicitly exposing reasoning logic, and must never directly reveal the core judgment point.
    \item The reasoning basis can fully utilize additional attribute information of each entity on the path (e.g., type, feature, scope) to enhance the complexity and concealment of the question.
    \item The implicit reasoning chain in the question must be \textbf{coherent}, simulating real human reasoning processes with natural logical connections (e.g., causal, conditional, associative), rather than mechanical splicing of entity-relationship pairs.
    \item The answer must be an exact and unique judgment (True/False) that strictly conforms to the logical consistency of the provided knowledge graph path or subgraph.
    \item The generated content must not contain redundant information such as "according to the knowledge graph", "based on a certain relationship", or explicit references to the knowledge graph itself.
    \item If the provided path or subgraph contains only one entity node, generate a reasonable and challenging judgment question based on the entity's core attributes (e.g., scope, function, characteristic) and ensure the answer is logically valid.
    \item The final output must be in strict JSON format, without any extra text outside the JSON structure.
\end{enumerate}

Please generate a question-answer pair that meets the requirements and return it in JSON format:
\begin{Verbatim}[breaklines=true, showspaces=false]
{
    "question": "Generated complex true/false question with natural contextual packaging",
    "answer": "Exact judgment (True/False)",
    "reasoning_path": "Coherent thinking process for question design, explaining how the question integrates multi-entity relationships/attributes, and why the answer is True/False—must be natural and in line with real reasoning logic"
}
\end{Verbatim}

}
\end{tcolorbox}

\subsection{Evaluation}
The following are the prompts for the three metrics used for evaluation.

\begin{tcolorbox}[breakable, title=Support, colframe=blue!75!black, colback=white, fonttitle=\bfseries, before skip=1pt, after skip=1pt, fontupper=\linespread{0.8}\selectfont]
{\footnotesize

\textbf{[Knowledge Support Assessment]}

Please evaluate strictly:
\begin{enumerate}
    \item Question: \{question\}
    \item Reference Answer: \{answer\}
    \item Chain of Thought: \{cot\} 
    \item Knowledge Graph Path: \{kgpath\}
\end{enumerate}

\textbf{Evaluation Criteria:}
\begin{itemize}
    \item Return 1 (Supported) if and only if:
    \begin{enumerate}
        \item The answer accurately addresses the question
        \item The reasoning is entirely based on the knowledge graph path
        \item No factual errors or hallucinations
    \end{enumerate}
    \item Return 0 (Not Supported) if any of the following exists:
    \begin{enumerate}
        \item The answer is incorrect or fails to address the question
        \item The reasoning is inconsistent with the knowledge graph path
        \item Contains information not present in the knowledge graph
    \end{enumerate}
\end{itemize}

\textbf{Note:} Only return the integer 0 or 1, without including any other content.

}
\end{tcolorbox}

\begin{tcolorbox}[breakable, title=Difficulty, colframe=blue!75!black, colback=white, fonttitle=\bfseries, before skip=1pt, after skip=1pt, fontupper=\linespread{0.8}\selectfont]
{\footnotesize

Now you are given a reading comprehension task. You will be provided with a question and corresponding reference information (support). You are required to answer the question based solely on the information given in the support and return the answer.

\textbf{Please note:}
\begin{enumerate}
    \item The answer should be derived directly from fragments of the support content. Do not include any extrapolated or additional information.
    \item Return the answer strictly in JSON format, for example:
    \{\{"answer":xxxx\}\}

\end{enumerate}

\noindent\textbf{Question:} \{question\}

\noindent\textbf{Support:} \{support\}

\noindent\textbf{Your answer:}

}
\end{tcolorbox}

\begin{tcolorbox}[breakable, title=Complexity, colframe=blue!75!black, colback=white, fonttitle=\bfseries, before skip=1pt, after skip=1pt, fontupper=\linespread{0.8}\selectfont]
{\footnotesize

You are a {professional complex question-answering data synthesis expert}, skilled at transforming {multi-hop relationships} in knowledge graphs into natural and fluent reasoning-based questions.

Please generate a high-quality multi-hop reasoning question-answer pair based on the provided knowledge graph subgraph or path information, along with relevant carrier evidence (images/tables/formulas, etc.).

\vspace{1em}
\noindent\textbf{Knowledge Graph Subgraph or Path Information:}

\{\{graph\_context\}\}

\begin{itemize}
    \item \textbf{Entity Nodes}: Real entities in the knowledge graph, such as people, locations, models, and tasks.
    \item \textbf{Carrier Nodes} (Table, Image, Figure, etc.): Used to provide relevant context for preceding entities and cannot be directly used as answers.
    \item \textbf{Relationships}: Relationships like \texttt{belong\_to} and \texttt{locate\_at} only indicate attribution or positioning and do not participate in the logical deduction of the reasoning chain.
\end{itemize}

\vspace{1em}
\noindent\textbf{Generation Requirements}
\begin{enumerate}
    \item \textbf{Question Design Principles}:
        \begin{itemize}
            \item The question must be derived by reasoning through relationships among multiple entities, reflecting the complexity of cross-entity, multi-hop logic. Avoid consecutive multiple questions.
            \item The question must be natural and fluent, avoiding mechanical phrasing and using natural language expression.
            \item The question must be contextually packaged rather than explicitly exposing reasoning steps. It is strictly prohibited to reveal the answer name or key words contained in the answer.
            \item If charts or images contain numerical values, trends, comparative relationships, etc., ensure that the generated question includes sufficient carrier-related context. Otherwise, the model cannot perform reasoning and answering. You need to incorporate enough information into the question as context to ensure the model can only derive the correct answer by "understanding the chart".
            \item \textbf{When the path includes image or table nodes}: Design reasoning types related to the carrier content, such as numerical comparison, trend changes, etc.
            \item The question should be based on the information presented by the carrier, but the final answer should still be the exact name of the entity connected to the carrier or its explicit attribute.
        \end{itemize}
    
    \item \textbf{Reasoning Process Requirements}:
        \begin{itemize}
            \item The reasoning chain can utilize additional attribute information of each entity on the path.
            \item The synthesized reasoning chain must be coherent, simulating real human reasoning processes with natural logical connectives. Do not explicitly label triple relationships; instead, write it as a coherent thinking process rather than mechanically listing entities and relationships.
            \item The generated content must not contain redundant information such as "according to the knowledge graph", "based on a certain relationship", "according to entity information", "the complete reasoning chain is as follows", or "according to a certain image/table".
            \item \textbf{When the endpoint is a chart/table/formula}: This node and its related attributes are only carrier evidence used to derive the exact name or specific attribute value of the \textbf{entity preceding the carrier}.
            \item Relationships like \texttt{belong\_to} and \texttt{locate\_at} only indicate such attribution relationships and are not used as reasoning relationships in the actual reasoning process. Analyze how the relevant carrier content supports or defines the preceding entity.
            \item When the path includes chart/image/table nodes, their content should be used as auxiliary evidence to support or limit the reasoning direction.
        \end{itemize}
    
    \item \textbf{Answer Requirements}:
        \begin{itemize}
            \item If the path endpoint contains carrier evidence such as "image/table/formula": The answer should be the exact name of the \textbf{entity preceding the carrier} (e.g., for path A$\rightarrow$B$\rightarrow$[Chart], the answer = B or its attribute value). Avoid overly long answers.
            \item It is strictly prohibited to use the carrier name or its related attributes as the answer.
            \item The answer must be unique, specific, and avoid ambiguity or polysemy.
            \item The answer should preferably not be numerical; it can be a trend change obtained through multi-step reasoning or the name of a relevant entity in the chart.
        \end{itemize}
    
    \item \textbf{Special Handling}:
        \begin{itemize}
            \item If the path contains only a single entity node:
                \begin{itemize}
                    \item For ordinary entities: Generate a question-answer pair based on its attributes.
                \end{itemize}
        \end{itemize}
\end{enumerate}

\vspace{1em}
\noindent\textbf{Output Specifications}
\begin{itemize}
    \item Return strictly in \textbf{JSON format} with no additional text (including markers like "```json").
    \item Required fields:
\end{itemize}

\begin{Verbatim}[breaklines=true, showspaces=false]
{
    "question": "A complex natural language question",
    "answer": "A concise, specific, and unique answer",
    "reasoning_path": "The reasoning process for question design, which is a CoT thinking or reasoning process that must be natural and coherent"
}
\end{Verbatim}

}
\end{tcolorbox}



\end{document}